\pdfoutput=1
\documentclass[journal]{IEEEtran}  

\usepackage{cite}
\usepackage[pdftex]{graphicx}
\usepackage{amsmath}
\usepackage{amssymb}  
\usepackage{algorithmicx}
\usepackage{mathtools}
\usepackage{url}
\usepackage[table]{xcolor} 
\usepackage{threeparttable}
\usepackage[breaklinks=true,bookmarks=false]{hyperref}
\hypersetup{
    colorlinks=true,
    linkcolor=blue,
    filecolor=magenta,      
    urlcolor=cyan,
}
\usepackage{caption}
\usepackage{subcaption}
\usepackage{makecell}
\usepackage{pifont}
\usepackage{float}

\title{\LARGE \bf
Driver Gaze Zone Estimation using Convolutional Neural Networks: \\A General Framework and Ablative Analysis
}

\author{Sourabh~Vora, Akshay~Rangesh, and Mohan~M.~Trivedi
\thanks{The authors are with the \href{http://cvrr.ucsd.edu/}{Laboratory for Intelligent and Safe Automobiles}, University of California, San Diego, CA 92092, USA.
\hfill \break
email - {\tt\small \{sovora,arangesh,mtrivedi\}@ucsd.edu}}
}

\begin{document}
\bstctlcite{IEEEexample:BSTcontrol}

\maketitle
\thispagestyle{empty}
\pagestyle{empty}

\begin{abstract}

Driver gaze has been shown to be an excellent surrogate for driver attention in intelligent vehicles. With the recent surge of highly autonomous vehicles, driver gaze can be useful for determining the \textit{handoff} time to a human driver. While there has been significant improvement in personalized driver gaze zone estimation systems, a generalized system which is invariant to different subjects, perspectives and scales is still lacking. We take a step towards this generalized system using Convolutional Neural Networks (CNNs). We finetune 4 popular CNN architectures for this task, and provide extensive comparisons of their outputs. We additionally experiment with different input image patches, and also examine how image size affects performance. For training and testing the networks, we collect a large naturalistic driving dataset comprising of 11 long drives, driven by 10 subjects in two different cars. Our best performing model achieves an accuracy of 95.18\% during \textit{cross-subject} testing, outperforming current state of the art techniques for this task. Finally, we evaluate our best performing model on the publicly available Columbia Gaze Dataset comprising of images from 56 subjects with varying head pose and gaze directions. Without any training, our model successfully encodes the different gaze directions on this diverse dataset, demonstrating good generalization capabilities.

\end{abstract}

\section{Introduction}

\IEEEPARstart{A}{ccording} to a recent study \cite{eriksson2016take} on `takeover time' in driverless cars, drivers engaged in secondary tasks exhibit larger variance and slower responses to requests to resume control. It is also well known that driver inattention is the leading cause of vehicular accidents. According to another study \cite{fitch2013impact}, 80\% of crashes and 65\% of near crashes involve driver distraction. 

Surveys on automotive collisions \cite{rueda2004influence, braitman2014effect} demonstrated that drivers were less likely (30\%-43\%) to cause an injury related collision when they had one or more passengers who could alert them to unseen hazards. It is therefore essential for Advanced Driver Assistance Systems (ADAS) to capture these distractions so that the driver can be alerted or guided in case of dangerous situations. This ensures that the handover process between the driver and the self driving car is smooth and safe.

Driver gaze is an important cue to recognize driver distraction. In a study on the effects of performing secondary tasks in a highly automated driving simulator \cite{li2016detecting}, it was found that the frequency and duration of mirror-checking reduced during secondary task performance versus normal, baseline driving. Alternatively, Ahlstrom et al. \cite{ahlstrom2013gaze} developed a rule based 2-second 'attention buffer' framework which depleted when the driver looked away from the field relevant to driving (FRD); and it starts filling up when the gaze direction is redirected towards FRD. Driver gaze activity can also be used to predict driver behavior \cite{doshi2011tactical}. Martin et al. \cite{martin2017behavior} developed a framework for modeling driver behavior and maneuver prediction from gaze fixations and transitions.

\begin{figure}[t!]
        \centering
        \begin{subfigure}[b]{0.475\columnwidth}
            \centering
            \includegraphics[width=\textwidth, height=3cm]{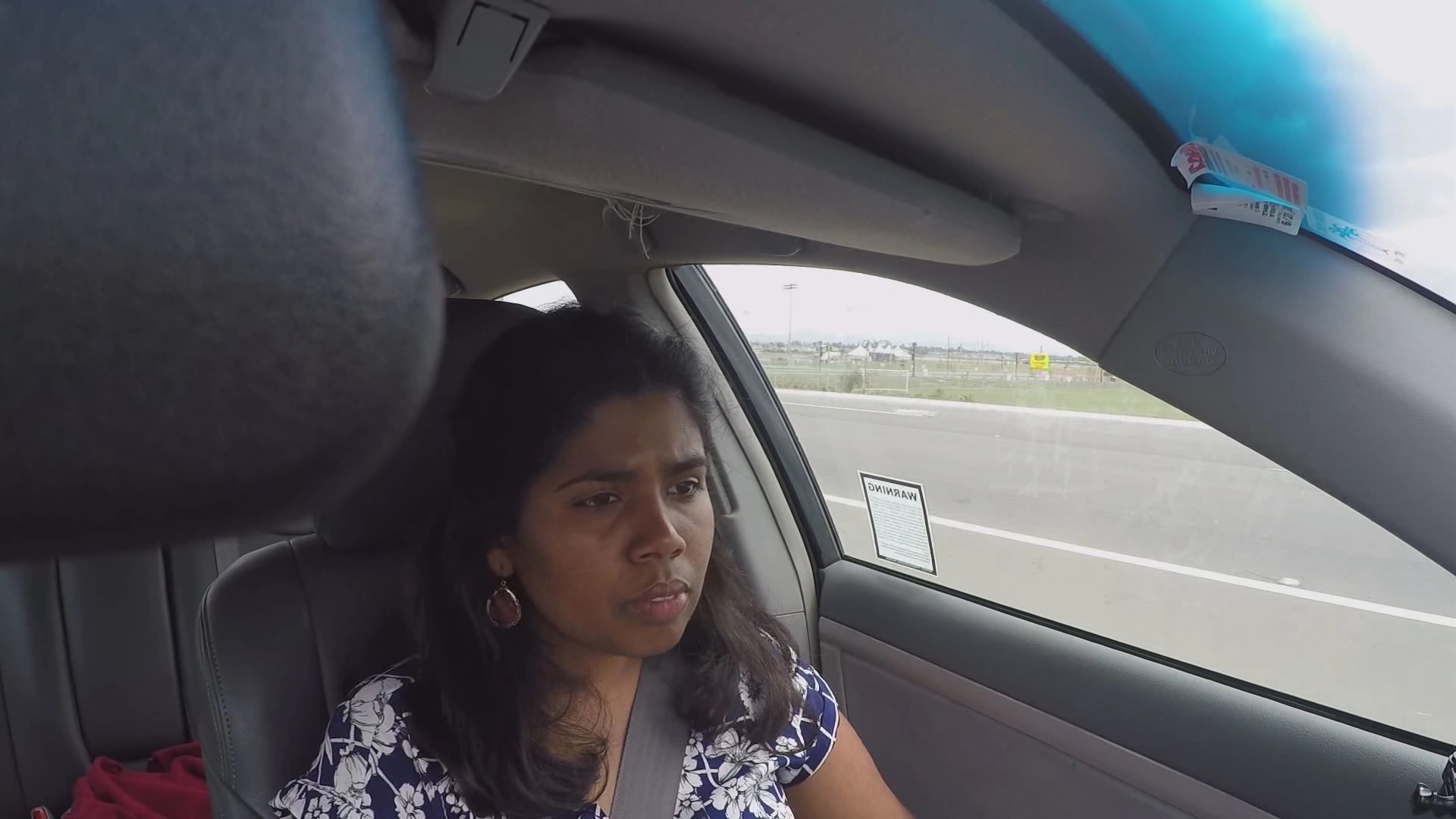}
        \end{subfigure}
        \hfill
        \begin{subfigure}[b]{0.475\columnwidth}  
            \centering 
            \includegraphics[width=\textwidth, height=3cm]{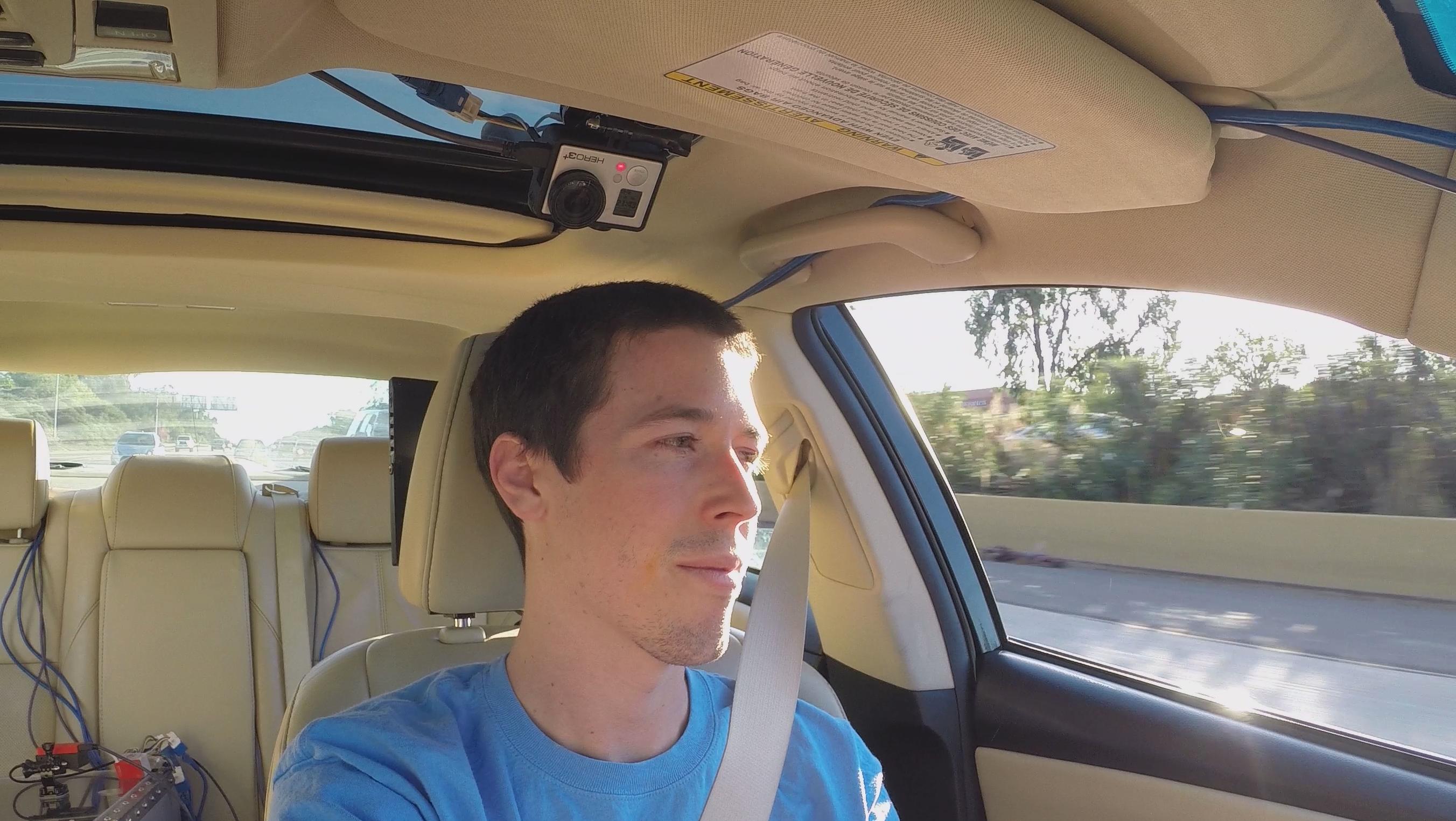}
        \end{subfigure}
        \vskip\baselineskip
        \begin{subfigure}[b]{0.475\columnwidth}   
            \centering 
            \includegraphics[width=\textwidth, height=3cm]{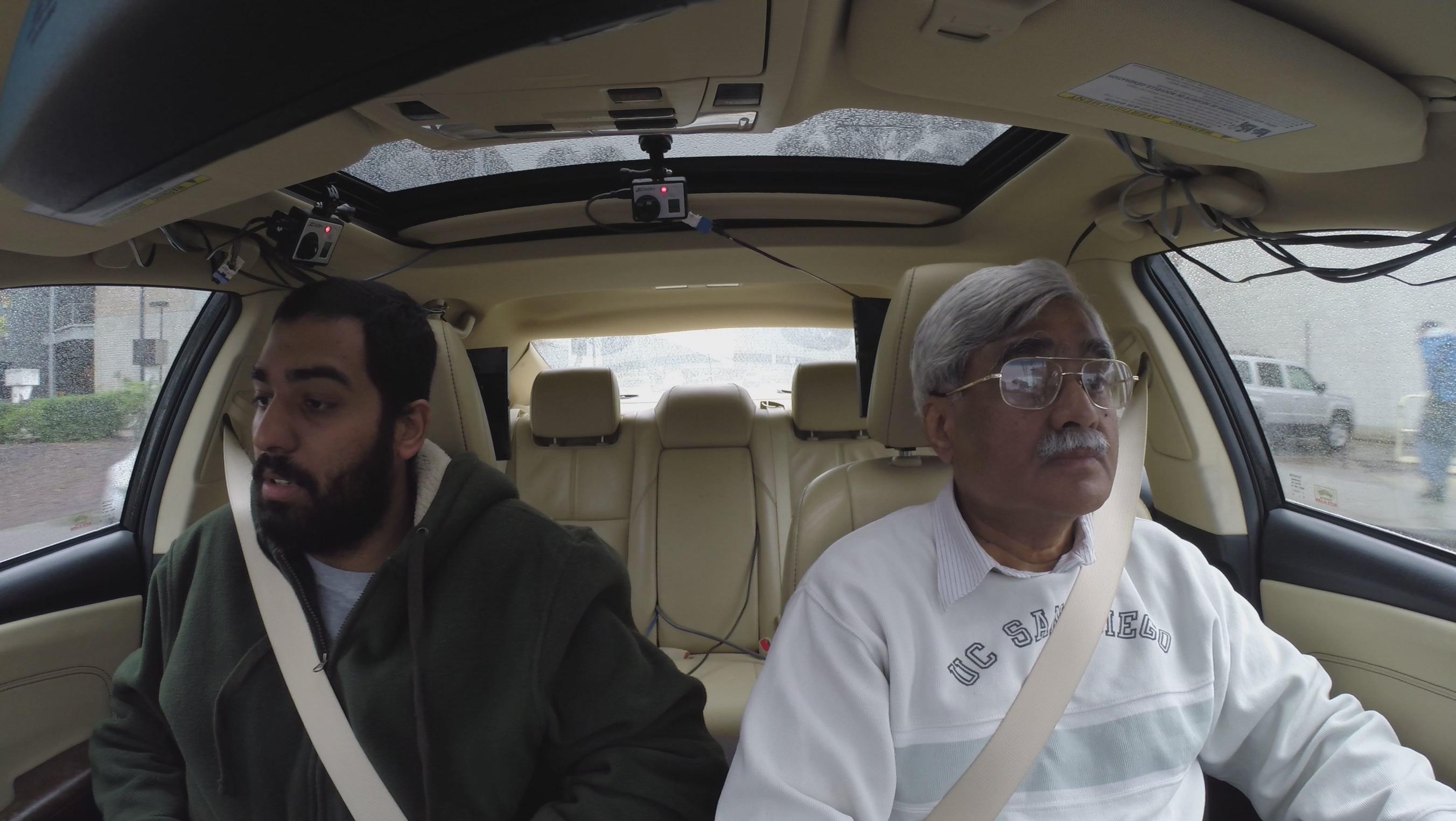}
        \end{subfigure}
        \hfill
        \begin{subfigure}[b]{0.475\columnwidth}   
            \centering 
            \includegraphics[width=\textwidth, height=3cm]{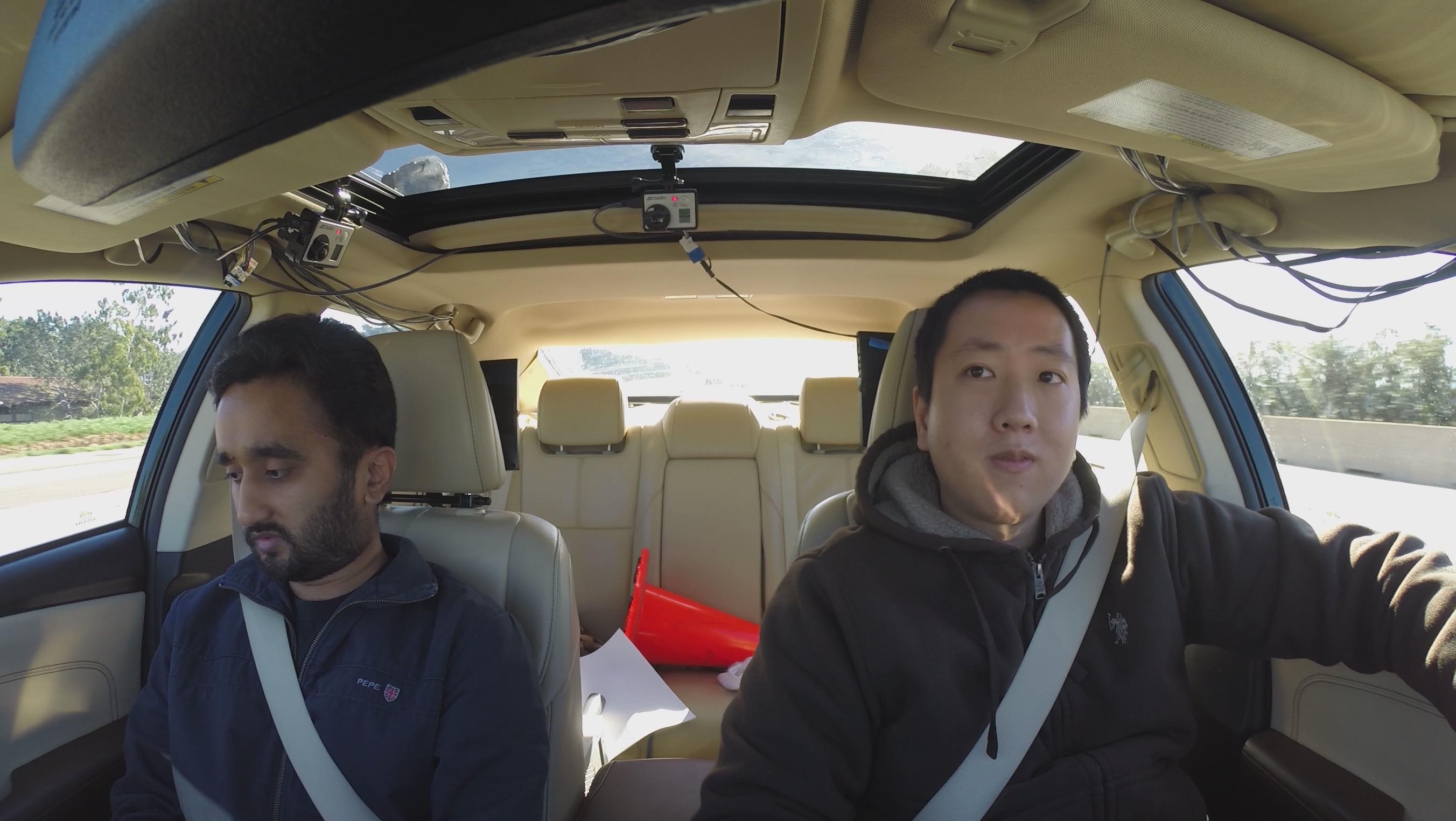}
        \end{subfigure}
        \caption{Where is the driver looking? Can a universal machine vision based system be trained to be invariant to drivers, perspective, scale, etc.?}
        \vspace{-3mm}
        \label{SampleImages}
\end{figure}

\begin{table*}[!ht]
\centering
\caption{Selected research studies on vision based driver gaze zone estimation systems in recent years.}
\label{comp}
\resizebox{\textwidth}{!}{%
\begin{tabular}{|c|c|c|c|c|c|l|}
\hline
\textbf{Research Study}                                          & \textbf{Objective}                                                                                                & \textbf{Camera}                                                    & \textbf{Features}                                                                                                                             & \textbf{\begin{tabular}[c]{@{}c@{}}Cross driver\\ testing\end{tabular}} & \textbf{\begin{tabular}[c]{@{}c@{}}Number\\ of\\ Zones\end{tabular}}             & \textbf{Classifier}                                                              \\ \hline\hline
\begin{tabular}[c]{@{}c@{}}Tawari and\\ Trivedi '14 \cite{tawari2014robust}\end{tabular} & \begin{tabular}[c]{@{}c@{}}Gaze zone estimation\\ using head pose dynamics\end{tabular}                           & \begin{tabular}[c]{@{}c@{}}2 cameras\\ with switching\end{tabular} & \begin{tabular}[c]{@{}c@{}}Head Pose static features\\ (yaw, pitch, roll),\\     Head Pose dynamic features\\ (6 per pose angle)\end{tabular} & \begin{tabular}[c]{@{}c@{}}No\end{tabular}      & 8                                                                                & \begin{tabular}[c]{@{}l@{}}Random\\ Forest\end{tabular}                      \\ \hline
\begin{tabular}[c]{@{}c@{}}Tawari et\\ al '14 \cite{tawari2014driver}\end{tabular}       & \begin{tabular}[c]{@{}c@{}}Gaze zone estimation\\ using head and eye cues\end{tabular}                            & \begin{tabular}[c]{@{}c@{}}2 cameras\\ with switching\end{tabular} & \begin{tabular}[c]{@{}c@{}}Head pose (yaw, pitch, roll),\\ Horizontal gaze,\\ Vertical gaze\end{tabular}                                      & \begin{tabular}[c]{@{}c@{}}No\end{tabular}      & 6                                                                                & \begin{tabular}[c]{@{}l@{}}Random\\ Forest\end{tabular}                      \\ \hline
\begin{tabular}[c]{@{}c@{}}Vasli et\\ al '16 \cite{vasli2016driver}\end{tabular}      & \begin{tabular}[c]{@{}c@{}}Gaze zone estimation\\ using fusion of geometric\\ and learning based method\end{tabular}                            & 1 Camera                                                           & \begin{tabular}[c]{@{}c@{}}Head Pose (yaw, pitch, roll),\\ 3d gaze, 2d - horizontal\\ and vertical gaze\end{tabular}                                                               & \begin{tabular}[c]{@{}c@{}}No\end{tabular}       & \begin{tabular}[c]{@{}c@{}}6\end{tabular} & \begin{tabular}[c]{@{}l@{}}SVM\end{tabular}                         \\ \hline
\begin{tabular}[c]{@{}c@{}}Fridman et\\ al '16 \cite{fridman2015driver}\end{tabular}      & \begin{tabular}[c]{@{}c@{}}Gaze zone estimation\\ using spatial configurations\\ of facial landmarks\end{tabular} & \begin{tabular}[c]{@{}c@{}}1 Camera\\ (Grayscale)\end{tabular}     & \begin{tabular}[c]{@{}c@{}}3 angles of each\\ triangles resulting from\\ Delaunay triangulation\\ over 19 facial landmarks\end{tabular}       & \begin{tabular}[c]{@{}c@{}}Yes\end{tabular}      & 6                                                                                & \begin{tabular}[c]{@{}l@{}}Random\\ Forest\end{tabular}                      \\ \hline
\begin{tabular}[c]{@{}c@{}}Fridman et\\ al '16 \cite{fridman2016owl}\end{tabular}      & \begin{tabular}[c]{@{}c@{}}Gaze zone estimation\\ using head and eye pose\end{tabular}                            & \begin{tabular}[c]{@{}c@{}}1 Camera\\ (Grayscale)\end{tabular}     & \begin{tabular}[c]{@{}c@{}}Head pose using nonlinear\\ classification of facial feature,\\ Pupil detection\end{tabular}                       & \begin{tabular}[c]{@{}c@{}}Yes\end{tabular}      & 6                                                                                & \begin{tabular}[c]{@{}l@{}}Random\\ Forest\end{tabular}                      \\ \hline
\begin{tabular}[c]{@{}c@{}}Choi et\\ al '16 \cite{choi2016real}\end{tabular}      & \begin{tabular}[c]{@{}c@{}}Gaze zone estimation\\ using CNN\end{tabular}                            & \begin{tabular}[c]{@{}c@{}}1 Camera\end{tabular}     & \begin{tabular}[c]{@{}c@{}}Automatically learned\\ using a Convolutional,\\ Neural Network\end{tabular}                       & \begin{tabular}[c]{@{}c@{}}No\end{tabular}      & 9                                                                                & \begin{tabular}[c]{@{}l@{}}Conv Net\end{tabular}                      \\ \hline
\begin{tabular}[c]{@{}c@{}}This study\end{tabular}        & \begin{tabular}[c]{@{}c@{}}Generalized Gaze zone \\estimation using CNNs\end{tabular}                            & 1 Camera                                                           & \begin{tabular}[c]{@{}c@{}}Automatically learned \\using a Convolutional\\ Neural Network\end{tabular}                            & \begin{tabular}[c]{@{}c@{}}Yes\end{tabular}      & 7                                                                                & \multicolumn{1}{c|}{\begin{tabular}[c]{@{}c@{}}Conv Net\end{tabular}} \\ \hline
\end{tabular}%
}
\end{table*}

While there has been a lot of research in improving personalized driver gaze zone estimation systems, there has been little progress in generalizing this task across different drivers, cars, perspectives and scale. We make an attempt in that direction using Convolutional Neural Networks (CNNs). CNNs have shown tremendous promise in the fields of image classification, object detection and recognition. CNNs are also good at transfer learning. Oquab et al. \cite{oquab2014learning} showed that image representations learned with CNNs on large-scale annotated datasets can be efficiently transferred to other visual recognition tasks. Therefore, instead of training a network from scratch, we adopt the transfer learning paradigm, where we finetune four different networks which have been trained to achieve state of the art results on the ImageNet \cite{deng2009imagenet} dataset. We analyze the effectiveness of each network in generalizing driver gaze zone estimation, by evaluating them on a large naturalistic driving dataset collected over 11 drives by 10 different subjects, in two different cars, each with slightly different camera settings and fields of view (Fig. \ref{SampleImages}).

The main contributions of this work are: a) A systematic ablative analysis of different CNN architectures and input strategies for generalizing driver gaze zone estimation systems b) Comparison of the CNN based model with some other state of the art approaches and, c) A large naturalistic driving dataset with extensive variability.

\section{Related Research}

Driver monitoring has been a long standing research problem in computer vision. For an overview on driver inattention monitoring systems, readers are encouraged to refer to a review by Dong et al. \cite{dong2011driver}. 

A prominent approach for driver gaze zone estimation is remote eye tracking. However, remote eye tracking is still a very challenging task in the outdoor environment. These systems \cite{bergasa2006real, ji2001real, ji2002real, morimoto2000pupil} rely on near-infrared (IR) illuminators to generate the bright pupil effect. This makes them sensitive to outdoor lighting conditions. Additionally, the hardware necessary to generate the bright eye effect hinders system integration. These specialized hardware also require a lengthy calibration procedure which is expensive to maintain due to constant vibrations and jolts experienced during driving. 

Owing to the above mentioned limitations, vision based systems appear to be an attractive solution for gaze zone estimation. These systems can be grouped into two categories: Techniques that only use the head pose \cite{tawari2014robust, lee2011real} and those that use the driver's head pose as well as gaze \cite{tawari2014driver, ishikawa2004passive, smith2003determining}. Driver head pose provides a decent estimate of the coarse gaze direction. For a good overview of vision based head pose estimation systems, readers are encouraged to refer to a survey by Murphy-Chutorian et al. \cite{murphy2009head}. However, methods which rely on head pose alone fail to discriminate between adjacent zones separated by subtle eye movement, like front windshield and speedometer. Tawari et al. \cite{tawari2014robust} combined static head pose with temporal dynamics in a multi-camera framework to obtain a more robust estimation of driver gaze. However, the problem of classifying driver gaze direction when he keeps his head static and uses only his eyes to look at different zones still persists.

It is therefore essential to 'look at' the driver's eyes. Tawari et al. \cite{tawari2014driver} combined head pose with the features extracted from facial landmarks on the eyes and achieved impressive results. Vasli et al. \cite{vasli2016driver} further used a fusion of head pose, features extracted from the eye as well as features obtained from the geometric constraints of the car to classify the driver's gaze into six zones. Fridman et al. \cite{fridman2016owl} also combined head pose and eye pose to classify driver gaze into 6 zones. The evaluations were commendably done on a large dataset comprising of 40 different drivers. 

There are two problems with the approaches described above: 1) Because they involve a complex pipeline of face detection, landmark estimation, pupil detection and finally feature extraction, the decision made by the classifier is completely dependent on the individual sub modules working correctly. 2) The hand crafted features designed from facial landmarks on the eyes are not completely robust to variations across different drivers, cars and seat positions. 

These problems come to light when the system is evaluated across variations like different subjects, cars, cameras and seat positions. To the best of our knowledge, the research studies by Fridman et al. \cite{fridman2016owl, fridman2015driver} are the only ones apart from ours that perform \textit{cross driver testing} (testing the system on drivers not seen during training) for the gaze zone estimation task. In their analysis on a huge dataset of 40 drivers, it was seen that in 40\% of the total annotated frames, the face or the pupil was not detected. Accurately detecting facial landmarks and pupils in real time under harsh illumination conditions inside a car is still a very challenging task, especially for profile faces. Further, they employ a high confidence decision pruning of 10 i.e. they only make a decision when the ratio of the highest probability predicted by the classifier to the second highest probability is greater than 10. This shows that their model does not generalize well to new drivers and overall, the decision making ability of their model is finally limited to 1.3 frames per second (fps) in a 30 fps video. A system with a low decision rate would miss several glances for mirror checks (a typical quick check of the rearview mirror or speedometer lasts less than a second). This would make such a system unusable for monitoring driver attention. 

A summary of recent studies on gaze zone estimation (involving 6 or more zones) using Naturalistic Driving Data (NDS) is shown in Table \ref{comp}. As can be seen, there are not many research studies on the effectiveness of CNNs for predicting the driver's gaze. Choi et al. \cite{choi2016real} use a five layered convolutional neural network to classify the driver's gaze into 9 zones. However, to the best of our knowledge, they do not conduct cross driver testing. In this study, we further systematize this approach by having separate subjects in the train and test sets. We also evaluate our model across variations in the camera position and field of view. This helps us test the generalization capability of CNNs for the gaze zone estimation task.

\begin{figure}[!t]
      \centering
      \includegraphics[width=\columnwidth]{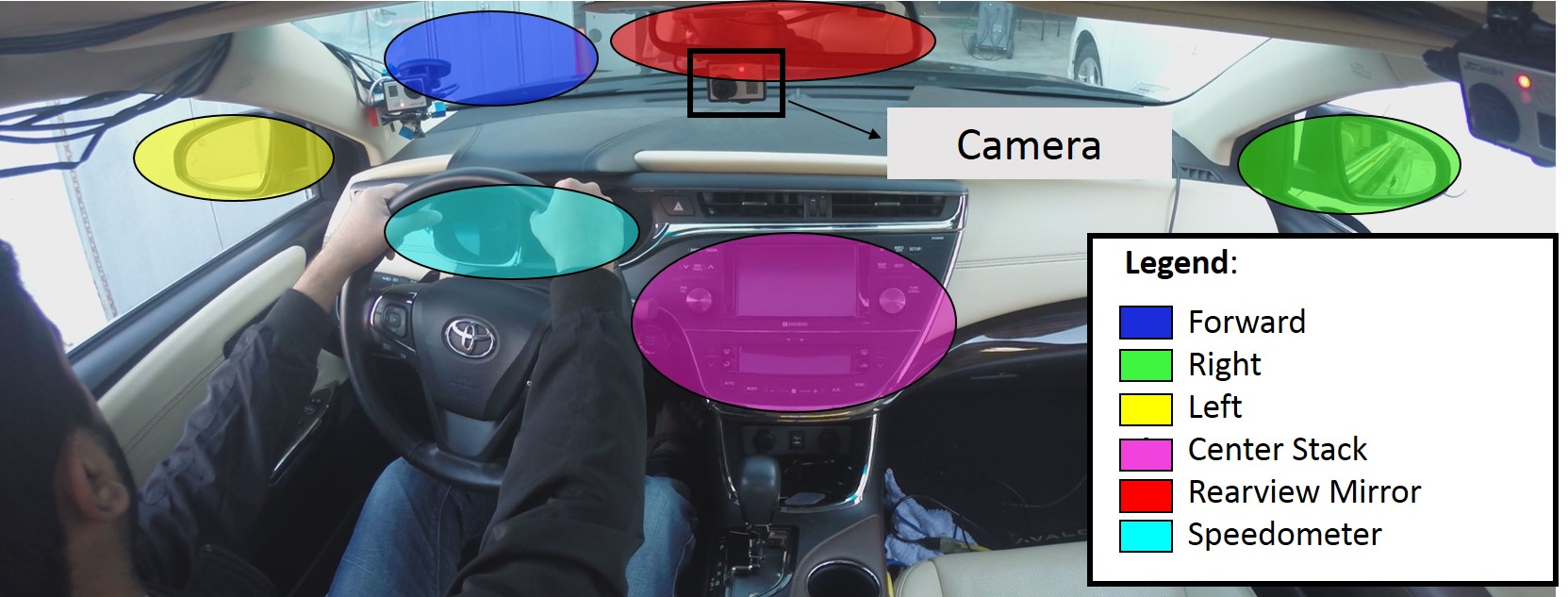}
      \caption{Illustration of the driver gaze zones considered in this study. We also highlight the approximate locations of the camera used to capture the input images.}
      \label{CarInterior}
\end{figure}

\section{Dataset}

Extensive naturalistic driving data was collected to enable us to train and evaluate our convolutional neural network models. Ten subjects drove two different cars instrumented with two inside looking cameras as well as one outside looking camera. The inside looking cameras capture the driver's face from different perspectives: one is mounted near the rear view mirror while the other is mounted near the A-pillar on the side window. The camera suite is time synchronized with all cameras capturing color video streams at 30 frames per second and a resolution of 2704 x 1524 pixels. The high resolution and the wide field of view captures both the driver and the passenger in a single frame.

While only images from the camera mounted near the rear-view mirror were used for our experiments, the other views were given to to a human expert for labeling the ground truth gaze zone. Seven different gaze zones (Fig. \ref{CarInterior}) are considered in our study- front windshield, right, left, center console (infotainment panel), center rear-view mirror, speedometer as well as an `eyes closed' state which usually occurs when the driver blinks.

11 different drives were recorded on different days and also at different times of the day. This was to ensure that our dataset contains sufficient variation in weather and consequently lighting. 10 different subjects participated in these drives. Table \ref{drives} describes the weather conditions for each drive and also lists the driver's age and gender.

\begin{table}[!t]
\centering
\caption{Dataset: Weather during the drive and driver's age and gender}
\label{drives}
\renewcommand{\arraystretch}{1.3}
\begin{tabular}{|c|c|c|c|c|}
\hline
\textbf{Drive} & \textbf{Weather} &\textbf{Time of drive} & \textbf{Driver's age} & \textbf{Gender} \\ \hline\hline
1             & Cloudy      &14:30 - 15:15			& 20-25		& Male                               \\ \hline
2             & Sunny       &16:30 - 17:05			& 25-30     & Male                               \\ \hline
3             & Sunny       &15:15 - 16:10			& 20-25		& Male                                \\ \hline
4  		      & Sunny       &13:45 - 14:15			& 20-25		& Female                                \\ \hline
5		      & Rainy       &12:10 - 13:20			& 60-65		& Male                                \\ \hline
6	          & Sunny       &17:10 - 17:40			& 20-25		& Female                                \\ \hline
7	          & Sunny       &12:20 - 12:50 			& 20-25		& Male									\\ \hline
8			  & Sunny       &16:05 - 16:30			& 20-25		& Male                                \\ \hline
9			  & Cloudy      &7:30 - 9:15 			& 25-30		& Female								\\ \hline
10			  & Sunny       &14:00 - 16:00 			& 25-30		& Female								\\ \hline
11			  & Cloudy		&11:45 - 12:30      	& 20-25		& Male									\\ \hline		
\end{tabular}
\end{table}

\begin{table}[!t]
\centering
\caption{Dataset: Number of annotated frames, frames used for training, and frames used for testing per gaze zone}
\label{frames}
\renewcommand{\arraystretch}{1.4}
\begin{tabular}{|c|c|c|c|}
\hline
\textbf{Gaze Zones} & \textbf{Annotated frames}	& \textbf{Training} & \textbf{Testing} \\ \hline\hline
Forward             & 21522 					& 3505              & 1023             \\ \hline
Right               & 4216 						& 3195              & 1021             \\ \hline
Left                & 4751						& 3725              & 1022             \\ \hline
Center Stack        & 4143						& 2831              & 1159             \\ \hline
Rearview Mirror     & 4489						& 3533              & 956              \\ \hline
Speedometer         & 4721						& 3580              & 1140             \\ \hline
Eyes Closed         & 3673						& 2565              & 1093             \\ \hline \hline
\textbf{Total}      & \textbf{47515}			& \textbf{22934}    & \textbf{7414}    \\ \hline
\end{tabular}
\end{table}

The frames for each zone were collected from a large number of 'events' separated well across time. An event is defined as a period of time in which the driver only looks at a particular zone. In a naturalistic drive, front facing events last for a longer time and also occur with highest frequency. Events corresponding to zones like Speedometer or Rearview Mirror usually last for a much smaller time and are sparser compared to front facing events. The objective of collecting the frames from a large number of events is to ensure sufficient variability in the head pose and pupil locations in the frames, as well as to obtain varied illumination conditions. Table \ref{frames} shows the distribution of the number of labeled frames per gaze zone.

Since forward facing frames dominate the dataset, they are sub-sampled to create a balanced dataset. Further, the dataset is divided such that drives from 7 subjects are used for training, while the drives from the remaining 3 subjects are used for testing to satisfy the cross subject testing requirement. This is particularly important as it helps us give an insight on whether the model generalizes well to different drivers. Table \ref{frames} shows the number of frames per zone finally used in our train and test datasets. The training set is further split into two subsets so as to create a validation set. We use a validation set comprising of 5\% of the training images. We ensured that the images of the training and validation set are not just different, but are also well separated in time. This is because frames captured at a particular time are very similar to each other. If we randomly divide the training set, we will end up having similar images in both training and validation sets which is not desirable.

Fig. \ref{SampleImages} shows some sample instances of drivers looking at different gaze zones. The videos were deliberately captured across different drives with different fields of view (wide angle vs normal). All subjects were also asked to adjust their seat positions according to their comfort. We believe that such variations in the dataset are necessary to build and evaluate a robust model that generalizes well.

\section{Proposed Method}

\begin{figure}[t!]
      \centering
      \includegraphics[width=0.5\textwidth]{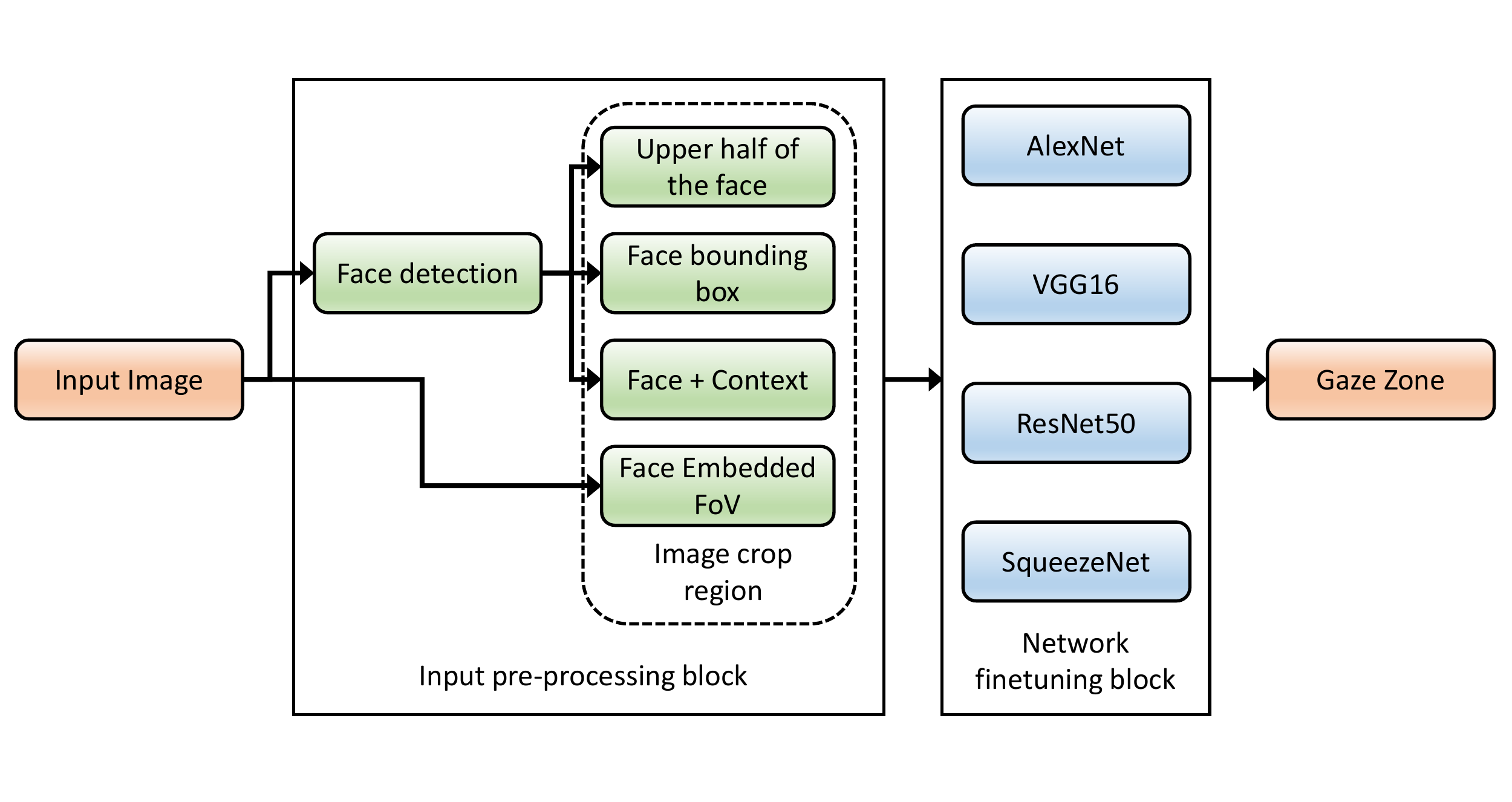}
      \caption{An overview of the proposed strategy for selecting the best performming CNN architecture and the best technique for pre-processing images, for the gaze zone estimation task. The whole process is divided into two blocks- the \textit{input pre-processing block}, and the \textit{network finetuning block}. Only one of the four input pre-processing technique and one of four CNN architectures are chosen during both training and testing.}
      \label{block}
\end{figure}

Fig. \ref{block} describes our strategy for selecting the best performing CNN architecture and the best technique for pre-processing images for the gaze zone estimation task. It consists of two major blocks, namely: a) Input pre-processing block and, b) Network finetuning block. The input pre-processing block extracts the sub image from the raw input image that is most relevant for gaze zone estimation. We consider four different pre-processing techniques. In the network finetuning block, we finetune four different CNNs using the sub images output by the input pre-processing block. 

Thus, we train 16 different CNNs, where each individual CNN was tuned on our validation set. We report the performance for each of the models (both accuracy and inference times) on the test set in Section \ref{exps}. Such ablation studies are very common in the recent literature\cite{he2016deep}, \cite{Simonyan14c} and can be used by a researcher to select a model based on their accuracy/runtime requirements. The following subsections describe the input pre-processing block, the network finetuning block and the training process in greater detail. 

\subsection{Network finetuning block}
\label{arch}
We finetune four CNNs which were originally trained on the ImageNet dataset \cite{deng2009imagenet}. We consider the following options: a) AlexNet, introduced by Krizhevsky et al. \cite{krizhevsky2012imagenet} b) VGG with 16 layers, introduced by Simonyan et al. \cite{Simonyan14c} c) ResNet with 50 layers, introduced by He et al. \cite{he2016deep} and d) SqueezeNet, introduced by Iandola et al. \cite{iandola2016squeezenet}. The motivation behind finetuning four different networks is to determine which network works best as well as to gain greater insights on the architectural details like depth, layers, kernel sizes and model sizes and how they affect the gaze zone classification task.

AlexNet is an eight layer CNN consisting of five convolution layers and two fully connected layers followed by a softmax layer. The first convolution layers have a large kernel size of $11\times11$ with a stride of 5, followed by $5\times5$ kernels in the 2nd layer and $3\times3$ kernels in the subsequent layers. VGG16 consists of 16 convolution and fully connected layers with a homogeneous architecture that only performs $3\times3$ convolutions and $2\times2$ pooling from the beginning to the end. Special skip connections were introduced in ResNet. It consists of $7\times7$ convolutions in the first layer followed by $3\times3$ kernels in the subsequent layers.  SqueezeNet consists of fire modules which are a special connection of $1\times1$ and $3\times3$ kernels. It has a very small model size and thus, the feasibility of FPGA and embedded deployment. Both Resnet50 and SqueezeNet have a gloabal average pooling layer at the end of the network. SqueezeNet follows up the global average pooling layer with the softmax non-linearity whereas Resnet50 includes a fully connected layer in between the pooling and softmax layers.

\subsection{Input pre-processing block}
\label{inp}

\begin{figure}
      	\centering
      	\includegraphics[width=0.5\textwidth]{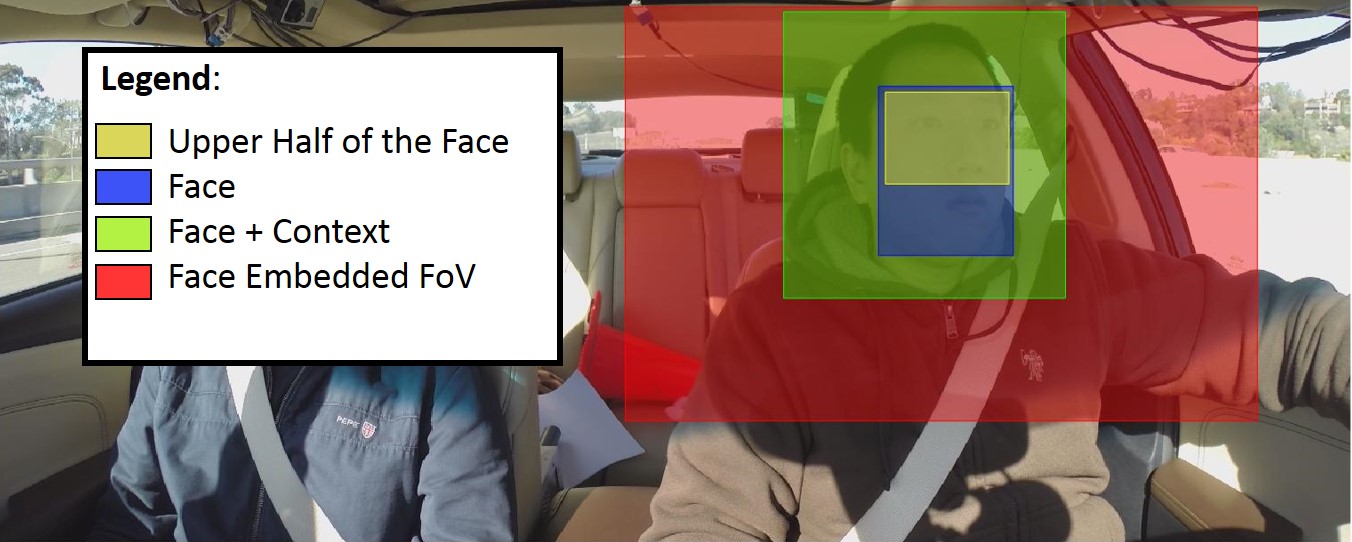}
        \caption{Different region crops on the input image that are used to train the CNNs. The crop regions are color coded for clarity.} 
        \label{InputImages}
\end{figure}

We choose four different approaches (Fig. \ref{InputImages}) for prepocessing the inputs to the CNNs while training. In the first case, driver's surround, which we call the Face-embedded field of view(FoV), was used as an input. This corresponds to the large sub image from the original image between the rearview mirror and (driver's) left rearview mirror. The head of the driver will always lie in this subimage. This will help us evaluate whether we can train our network directly from the input images, thereby skipping the face detection step. In the second case, driver's face was detected and used as an input to the CNNs. The face detector presented by Yuen et al. \cite{yuen2016looking} was used in our experiments. In the third pre processing strategy, some context was added to driver's face by extending the face bounding box in all directions. The thought process behind adding context to the driver's face is to learn features which determine the position of the driver's head with respect to his fixed surroundings. Adding context has given a boost in performance in several computer vision problems and this input strategy will help us determine whether it's the same for the driver gaze zone classification task. In the fourth pre-processing approach, only the top half of the face was used as an input. The extracted images were all resized to 224x224 or 227x227 according to the network requirements and finally, the mean was subtracted. 

\subsection{Training}
\label{training}

For AlexNet and VGG16 and Resnet50 architectures, we replace the last layer of the network (which has 1000 neurons) with a new fully connected layer with 7 neurons and add a softmax layer on top of it. For SqueezeNet, we limit the number of kernels in the last convolution layer from 1000 to 7. We initialize the newly added layers using the method proposed by He et al. \cite{he2015delving}. We finetune the entire network using our training data. Since the networks are already pre-trained on a very large dataset, we use a low learning rate. For all networks, we start with a hundredth of the learning rate used to train the respective networks and observe the training and validation loss and accuracy. If the loss function oscillates, we further decrease the learning rate. It was found that a learning rate of $4 \times 10^{-4}$ works well with SqueezeNet while a learning rate of $10^{-4}$ works well with the other three networks. All the networks were finetuned for a duration of 50 epochs with mini batch gradient descent using adaptive learning rates. Beyond 50 epochs, the networks started to overfit. Based on GPU memory constraints, batch sizes of 64, 64, 32 and 16 were used for training AlexNet, SqueezeNet, VGG16 and ResNet50 respectively. The Adam optimization algorithm, introduced by Kingma and Ba \cite{kingma2014adam}, was used. Data augmentation by flipping or rotating the images wasn't performed as it can either potentially change the labels of the image or generate unrealistic images which won't be seen during normal driving. Changing the pixel intensities was possible but we decided to go against it because our dataset already had extensive variation in illumination. All experiments were performed on the Caffe \cite{jia2014caffe} framework.

\section{Experimental Analysis \& Discussion}
\label{exps}

The evaluation of the experiments performed in IV are presented using three metrics. The first two forms of evaluation metrics are the macro-average and micro-average accuracy. They are calculated as:
\begin{equation}
  \text{Macro-average accuracy}= \frac{1}{N} \sum_{i=1}^N \frac{(\text{True positive})_i}{(\text{Total Population})_i}
\end{equation}

\begin{equation}
  \text{Micro-average accuracy}= \frac{\sum_{i=1}^N (\text{True positive})_i}{\sum_{i=1}^N (\text{Total Population})_i}
\end{equation}

where, N $=$ Number of gaze zones.\\
The third evaluation metric is the N class confusion matrix where each row represents true gaze zone and each column represents estimated gaze zone. 

The face detector used in our experiments \cite{yuen2016looking} is currently the best performing face detector on the VIVA-Face dataset \cite{martin2016vision}, which comprises of images sampled from 39 naturalistic driving videos, featuring harsh lighting conditions and facial occlusions. For a detailed analysis of its performance, readers are advised to refer to \cite{yuen2016looking}. We observed less than 0.25\% false detections on our training set. As it is very robust, we don't check for false detections on our test set and the performance reported in the following sections will therefore be the true performance of our system.

\subsection{Analysis of network architectures and different image crop regions}
\label{nets_ana}

Table \ref{acc_table} presents the macro-average accuracy obtained on the test set for sixteen different combinations of networks and image crop regions. Two trends are clearly observable from Table \ref{acc_table}. First, the performance of all three networks improves as the network is provided a higher resolution image of the eye while training and testing. It can be seen that all the networks perform best when only the upper half of the face is given as an input to the network. Second, the SqueezeNet architecture consistently outperforms VGG16 which further outperforms ResNet50 for all different image crop regions. AlexNet does not do as well as compared to the other three networks, particularly when the eyes of the driver are a very small part of the image. Our best performing model is a finetuned SqueezeNet trained on the images of the upper half of the face , which achieves an accuracy of 95.18\% and clearly demonstrates the generalization capabilities of the features learned through CNN.

\begin{figure}[!t]
        \centering
        \begin{subfigure}[b]{0.3\columnwidth}
            \centering
            \includegraphics[width=2.7cm]{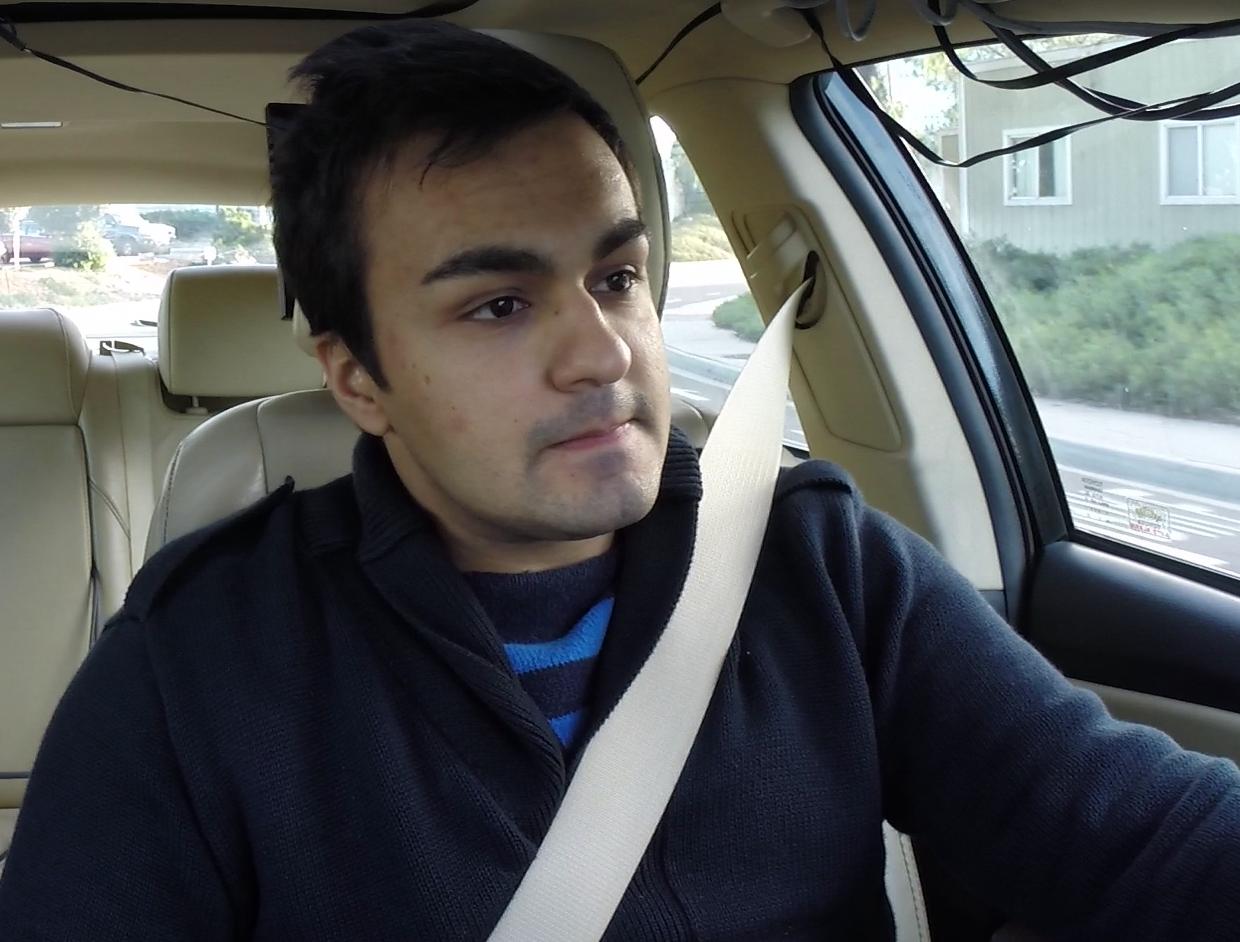}
            \caption{\small Forward}
        \end{subfigure}
        \begin{subfigure}[b]{0.3\columnwidth}  
            \centering 
            \includegraphics[width=2.7cm]{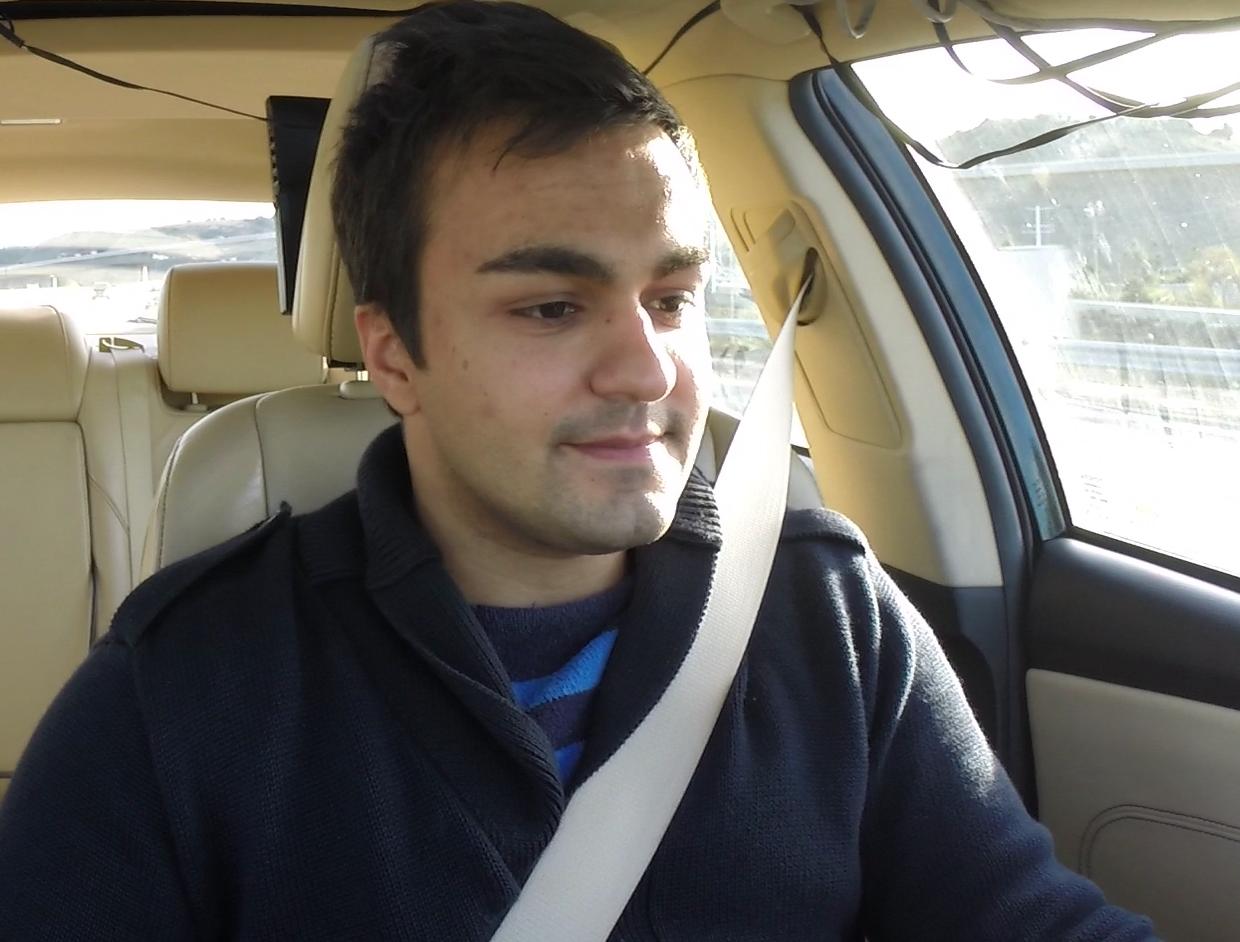}
            \caption{\small Speedometer}
        \end{subfigure}
        \begin{subfigure}[b]{0.3\columnwidth}   
            \centering 
            \includegraphics[width=2.7cm]{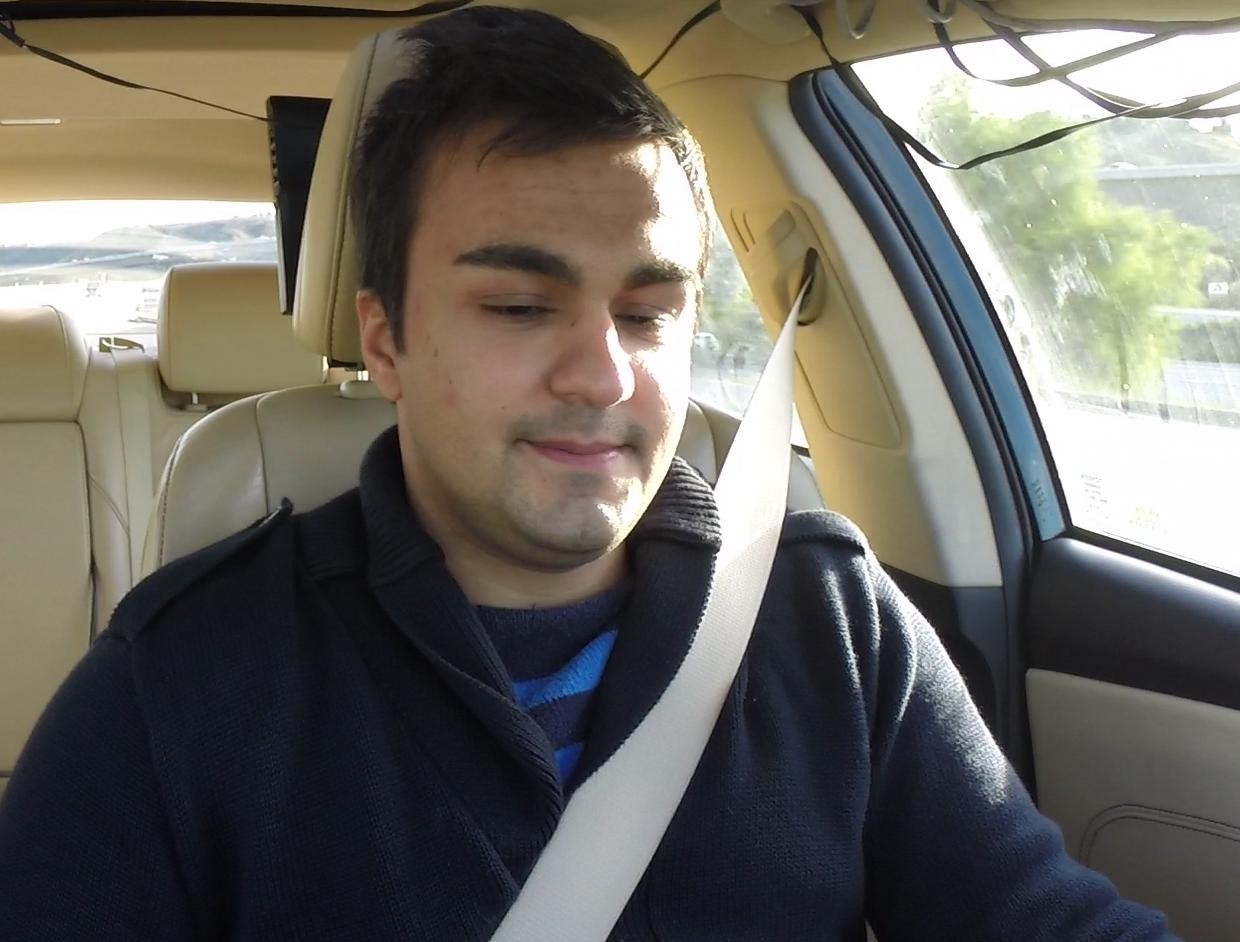}
            \caption{\small Eyes closed}
        \end{subfigure}
        \caption{Example image that illustrates the subtle differences in the eye when the driver is looking at three different zones.} 
        \vspace{-4mm}
        \label{eye_diff}
\end{figure}

It is particularly interesting to note the very low performance of finetuned AlexNet when using the Face-embedded FoV images as compared to the other three networks. This can be attributed to the large kernel size ($11\times11$) and a stride of 4 in the first convolution layer. The gaze zones change with very slight movement of the pupil or eyelid. We feel that this fine discriminating information of the eye is missed out in the first few layers due to large convolution kernels and large strides. In our experiments, we found that the network easily classifies zones with large head movement (left and right) whereas it struggles to classify zones with slight eye movement (Eg. Front, Speedometer and Eyes Closed (Fig. \ref{eye_diff})). The large increase in accuracy when only the top half of the face is provided as an input as compared to when the large sub image is provided further confirms the fact. This dependence on the resolution of the eye seen by the network is further elaborated upon in \ref{noface}.

SqueezeNet consists of a combination of $3\times3$ and $1\times1$ kernels while VGG16 is composed of convolution layers that perform $3\times3$ convolutions with a stride of 1. These small convolution kernels coupled with the larger depth of the network allows for learning features which help to discriminate gaze zones with even slight movements of the pupil or eyelid. This enable them to perform much better than AlexNet.

With ResNet50, we consistently achieve a slightly lower accuracy on the test set as compared to SqueezeNet and VGG16 for all input pre-processing approaches. This could be again because of the large convolution kernel in the first layer (7$\times$7). Another possible reason can be the limited amount of training data to fine tune a much deeper (50-layered) network.

The results in the form of confusion matrices and accuracies, when the networks were trained for half face images, are further shown in Tables \ref{OurModel}, \ref{vgg}, \ref{alexnet} and \ref{resnet} for finetuned SqueezeNet, VGG16, AlexNet and Resnet50 respectively. 

\begin{table}[!t]
\centering
\caption{Ablation experiments with different CNNs and different image crop regions. Macro-average accuracy obtained for each experiment is tabulated.}
\renewcommand{\arraystretch}{2.2}
\resizebox{\columnwidth}{!}{%
\begin{tabular}{|c|c|c|c|c|}
\hline
        \textbf{Architecture}    & \textbf{Half Face} & \textbf{Face}  & \textbf{Face+Context} & \textbf{\begin{tabular}[c]{@{}c@{}}Face Embedded\\ FoV\end{tabular}} \\ \hline\hline
{AlexNet}  & 88.91             & 82.08         & 75.56               & 62.21              \\ \hline
{ResNet50} & 91.66               & 89.34        & 86.67                 & 87.04                \\ \hline
{VGG16}    & 93.36      & 92.74 				& 91.21      & 88.92     \\ \hline
{SqueezeNet} & \textbf{95.18}		& \textbf{94.81}			& \textbf{92.74}		& \textbf{89.37}  \\ \hline
\end{tabular}
}
\label{acc_table}
\end{table}

\subsection{Comparison of our CNN based model with some current state of the art models}

In this section, we compare our best performing model (SqueezeNet trained on upper half of face images) with some other recent gaze zone estimation studies. The technique presented by Tawari et al. \cite{tawari2014driver} was implemented on our dataset so as to enable a fair comparison. They use a Random Forest classifier with hand crafted features of head pose and gaze surrogates which are computed using facial landmarks.

Table \ref{OurModel} presents the confusion matrix obtained by testing our finetuned SqueezeNet model while Table \ref{OtherModel} presents the confusion matrix obtained by the Random Forest Model. We see that our CNN based model clearly outperforms the Random Forest model by a substantial margin of 26.42\%. There are several factors responsible for the low performance of the Random Forest model. The Random Forest model uses head pose and gaze angles as the features to discriminate between different gaze zones and these angles are not robust to the position and orientation of the driver with respect to the camera. This problem is further highlighted in our dataset because it consists of images captured under different settings of field of view. The angle measures are further distorted because of incorrect landmark estimation particularly for profile or partially occluded faces. Further, for determining the eye openness, the area of the upper eyelid is used in the Random Forest model. Eye area is again not a robust feature as it changes with different subjects, different seat position and different camera settings. All these factors combined limit the Random Forest model to generalize, as shown by the results on our dataset. 

We also compare our work with Choi et al. \cite{choi2016real}, who used a truncated version of AlexNet and achieved a high accuracy of 95\% on their own dataset. However, to the best of our knowledge, they don't do cross driver testing and divide each drive temporally. The first 70\% frames for each drive were used for training, next 15\% frames were used for validation while the last 15\% were used for testing. In our experiments (Table \ref{acc_table}), we show that AlexNet does not perform very well as compared to the other networks considered by us. When we tried to replicate their experimental setup by dividing each drive temporally (thereby training and testing on the images of same drivers) and using the resized face images as input to our network, we achieve a very high accuracy of 98.7\%. When tested on different drivers, the accuracy drops down substantially to 82.5\%. This clearly shows that the network is over fitting the task by learning driver specific features.

\begin{table}[!t]
\begin{center}
\caption{Confusion matrix for 7 gaze zones using finetuned SqueezeNet trained on images containing upper half of the face.}
\label{OurModel}
\renewcommand{\arraystretch}{2.4}
\resizebox{\columnwidth}{!}{%
\begin{tabular}{|c|c|c|c|c|c|c|c|}
\hline
\textbf{True Zone} & \multicolumn{7}{c|}{\textbf{Recognized Gaze Zone}}                                                                            \\ \cline{1-8} 
Forward & \textbf{97.65}   & 0    & 1.17         & 0              & 0.68           & 0.39           & 0.1            \\ \hline
Right & 0    & \textbf{100}     & 0            & 0              & 0            & 0              & 0           \\ \hline
Left & 3.23     & 0          & \textbf{94.03} & 0              & 0              & 0.1            & 2.64          \\ \hline
Center Stack & 0.09     & 7.77         & 0            & \textbf{90.42} & 0              & 1.12           & 0.6           \\ \hline
Rearview Mirror & 0        & 0.1          & 0            & 0         & \textbf{99.9}       & 0              & 0.31           \\ \hline
Speedometer & 5.79     & 0            & 0.09         & 2.63           & 0              & \textbf{89.21} & 2.28           \\ \hline
Eyes Closed & 0.73     & 0.37         & 1.83         & 1.28           & 0              & 0.73            & \textbf{95.06} \\ \hline
\end{tabular}
}
\end{center}
Macro-average Accuracy = 95.18\%
\\Micro-average Accuracy = 94.96\%
\end{table}

\begin{table}[!t]
\begin{center}
\caption{Confusion matrix for 7 gaze zones using the Random Forest model.}
\label{OtherModel}
\renewcommand{\arraystretch}{2.4}
\resizebox{\columnwidth}{!}{%
\begin{tabular}{|c|c|c|c|c|c|c|c|}
\hline
\textbf{True Zone} & \multicolumn{7}{c|}{\textbf{Recognized Gaze Zone}}                                                                            \\ \cline{1-8} 
Forward                                                                           & \textbf{84.16} & 0              & 7.72           & 0.68           & 1.47           & 5.38           & 0.59           \\ \hline
Right                                                                           & 0              & \textbf{99.12} & 0              & 0              & 0.39           & 0              & 0.49           \\ \hline
Left                                                                           & 6.17           & 0              & \textbf{71.17} & 0.33           & 0.67           & 1.83           & 19.83          \\ \hline
Center Stack                                                                           & 0.78           & 8.57           & 0              & \textbf{32.55} & 15.41          & 0              & 42.68          \\ \hline
Rearview Mirror                                                                           & 0              & 0.84           & 0              & 0.21           & \textbf{98.74} & 0              & 0.21           \\ \hline
Speedometer                                                                           & 27.81          & 0              & 6.84           & 2.89           & 0              & \textbf{40.96} & 21.49          \\ \hline
Eyes Closed                                                                           & 6.99           & 4.56           & 11.36          & 10.87          & 7.18           & 4.37           & \textbf{54.66} \\ \hline
\end{tabular}
}
\end{center}
Macro-average Accuracy = 68.76\%
\\Micro-average Accuracy = 67.15\%
\end{table}

\begin{table}[!t]
\begin{center}
\caption{Confusion matrix for 7 gaze zones using finetuned VGG16 trained on images containing upper half of the face.}
\label{vgg}
\renewcommand{\arraystretch}{2.4}
\resizebox{\columnwidth}{!}{%
\begin{tabular}{|c|c|c|c|c|c|c|c|}
\hline
\textbf{True Zone}     & \multicolumn{7}{c|}{\textbf{Recognized Gaze Zone}}                                                                            \\ \hline
Forward         & \textbf{95.31} & 0.2            & 1.56           & 0              & 1.76           & 1.08           & 0.1            \\ \hline
Right           & 0              & \textbf{99.51} & 0              & 0              & 0.1            & 0              & 0.39           \\ \hline
Left            & 1.96           & 0              & \textbf{85.71} & 0              & 0              & 0.1            & 12.23          \\ \hline
Center Stack    & 0.17           & 6.56           & 0.35           & \textbf{87.58} & 0.26           & 4.92           & 0.17           \\ \hline
Rearview Mirror & 0              & 0.21           & 0              & 0              & \textbf{99.48} & 0              & 0.31           \\ \hline
Speedometer     & 1.49           & 0.61           & 0              & 5.27           & 0              & \textbf{90.87} & 1.76           \\ \hline
Eyes Closed     & 0.91           & 0.82           & 0.18           & 0.73           & 0.09           & 2.2            & \textbf{95.06} \\ \hline
\end{tabular}
}
\end{center}
Macro-average Accuracy = 93.59\%
\\Micro-average Accuracy = 93.17\%
\end{table}

\begin{table}[!t]
\begin{center}
\caption{Confusion matrix for 7 gaze zones using finetuned AlexNet trained on images containing upper half of the face.}
\label{alexnet}
\renewcommand{\arraystretch}{2.4}
\resizebox{\columnwidth}{!}{%
\begin{tabular}{|c|c|c|c|c|c|c|c|}
\hline
\textbf{True Zone}       & \multicolumn{7}{c|}{\textbf{Recognized Gaze Zone}}                                                                            \\ \hline
Forward         & \textbf{85.92} & 0.29          & 2.05           & 0              & 9.68           & 1.86           & 0.2            \\ \hline
Right           & 0              & \textbf{99.9} & 0              & 0              & 0              & 0              & 0.1            \\ \hline
Left            & 1.57           & 0             & \textbf{84.54} & 0.39           & 0              & 1.86           & 11.64          \\ \hline
Center Stack    & 0              & 9.92          & 0              & \textbf{74.55} & 0.52           & 3.28           & 11.73          \\ \hline
Rearview Mirror & 0              & 1.36          & 0              & 0              & \textbf{98.64} & 0              & 0              \\ \hline
Speedometer     & 5.61           & 0             & 1.32           & 4.21           & 0.09           & \textbf{86.49} & 2.28           \\ \hline
Eyes Closed     & 3.39           & 0.73          & 0.64           & 1.65           & 0.55           & 0.73           & \textbf{92.31} \\ \hline
\end{tabular}
}
\end{center}
Macro-average Accuracy = 88.55\%
\\Micro-average Accuracy = 88.91\%
\end{table}

\begin{table}[!t]
\begin{center}
\caption{Confusion matrix for 7 gaze zones using finetuned ResNet50 trained on images containing upper half of the face.}
\label{resnet}
\renewcommand{\arraystretch}{2.4}
\resizebox{\columnwidth}{!}{%
\begin{tabular}{|c|c|c|c|c|c|c|c|}
\hline
\textbf{True Zone}       & \multicolumn{7}{c|}{\textbf{Recognized Gaze Zone}}                                                                          \\ \hline
Forward         & \textbf{86.12} & 0.1            & 2.15           & 0              & 9.68         & 0.49           & 1.47           \\ \hline
Right           & 0              & \textbf{96.67} & 0              & 0.1            & 0            & 0              & 3.23           \\ \hline
Left            & 1.17           & 0              & \textbf{90.22} & 0              & 0.2          & 0.1            & 8.32           \\ \hline
Center Stack    & 0.26           & 6.21           & 0              & \textbf{89.99} & 0.26         & 0              & 3.28           \\ \hline
Rearview Mirror & 0              & 0              & 0              & 0              & \textbf{100} & 0              & 0              \\ \hline
Speedometer     & 1.49           & 0              & 0.35           & 3.6            & 0            & \textbf{79.37} & 15.19          \\ \hline
Eyes Closed     & 0.09           & 0.18           & 0.09           & 0.37           & 0            & 0              & \textbf{99.27} \\ \hline
\end{tabular}
}
\end{center}
Macro-average Accuracy = 91.43\%
\\Micro-average Accuracy = 91.66\%
\end{table}

\subsection{How can we get away without face detection?}
\label{noface}

In \ref{nets_ana}, we observed that the finetuned SqueezeNet model performs very well (Table \ref{acc_table}) even on Face-embedded FoV images. In fact, all finetuned network architectures apart from AlexNet perform well. In this section, we attempt to understand what the network is learning and determine whether it is able to focus on driver's eyes, which is such a small part of the image.

We consider the finetuned SqueezeNet model for the experiments in this section as it was shown to perform the best in \ref{nets_ana}. In the SqueezeNet architecture, there are no fully connected layers. The final convolution layer has seven filters producing seven class activation maps (CAMs) which correspond to the seven gaze zones considered in this research. The final convolution layer is followed by the global average pooling (GAP) layer and finally the softmax layer. Zhou et al. \cite{zhou2016learning} showed that the GAP layer explicitly enables the CNN to have remarkable localization ability despite being trained on  image level labels. We see this further in our experiments.  We consider three sample images (Image A, Image B and Image C) and visualize the seven class activation maps (CAMs) obtained before the GAP layer. We generate these CAMs when the SqueezeNet model was finetuned on different image crop regions i.e. upper half of the face, face bounding box, face and context and Face-embedded FoV. The generated CAMs were resized to the size of the image ($224\times224$) so as to enable us to see where the activations localize on the image. 

\begin{figure*}[t!]
      \centering
      \includegraphics[width=0.9\linewidth, height=22cm]{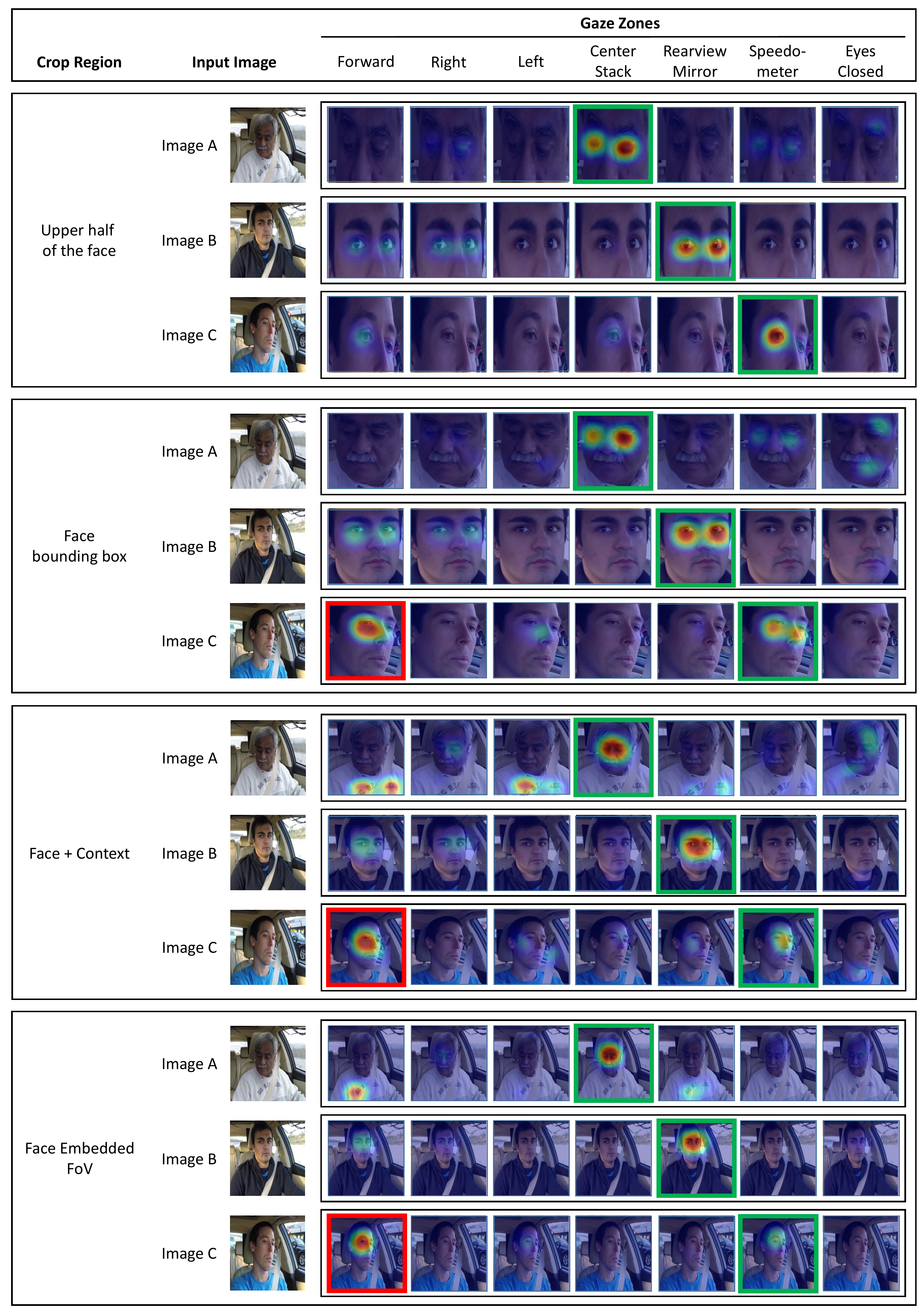}
      \caption{Class activation maps (CAMs) for seven gaze zones considered in this research for three sample images (A, B and C). The four major rows correspond to the image region crops on which the network was trained on. The green boxes shows the ground truth class labels while the red boxes shows if the network made an incorrect prediction. It can be observed that our model learns to localize the eyes of the driver. This is true even when no bounding box labels of the eyes or the face was provided to the network when it was trained on driver vicinity images.}
      \label{bigImg}
\end{figure*}

Fig. \ref{bigImg} visualizes all the CAMs. It is composed of four major rows where each major row corresponds to the networks trained on different image crop regions. Each major row is further subdivided into three sub rows, where each sub row corresponds to the activations visualized over the image crop regions of the original test image. We gain several insights from visualizing the CAMs. 

First, the activations always localize over the eyes of the driver. This is true even when the network was trained on Face-embedded FoV images where the eyes form a really small part of the image. This is particularly fascinating since the network was not provided any bounding box labels of the eyes or the face and it has learned to effectively localize the eyes. 

\begin{figure}[t!]
      \centering
      \includegraphics[width=\linewidth]{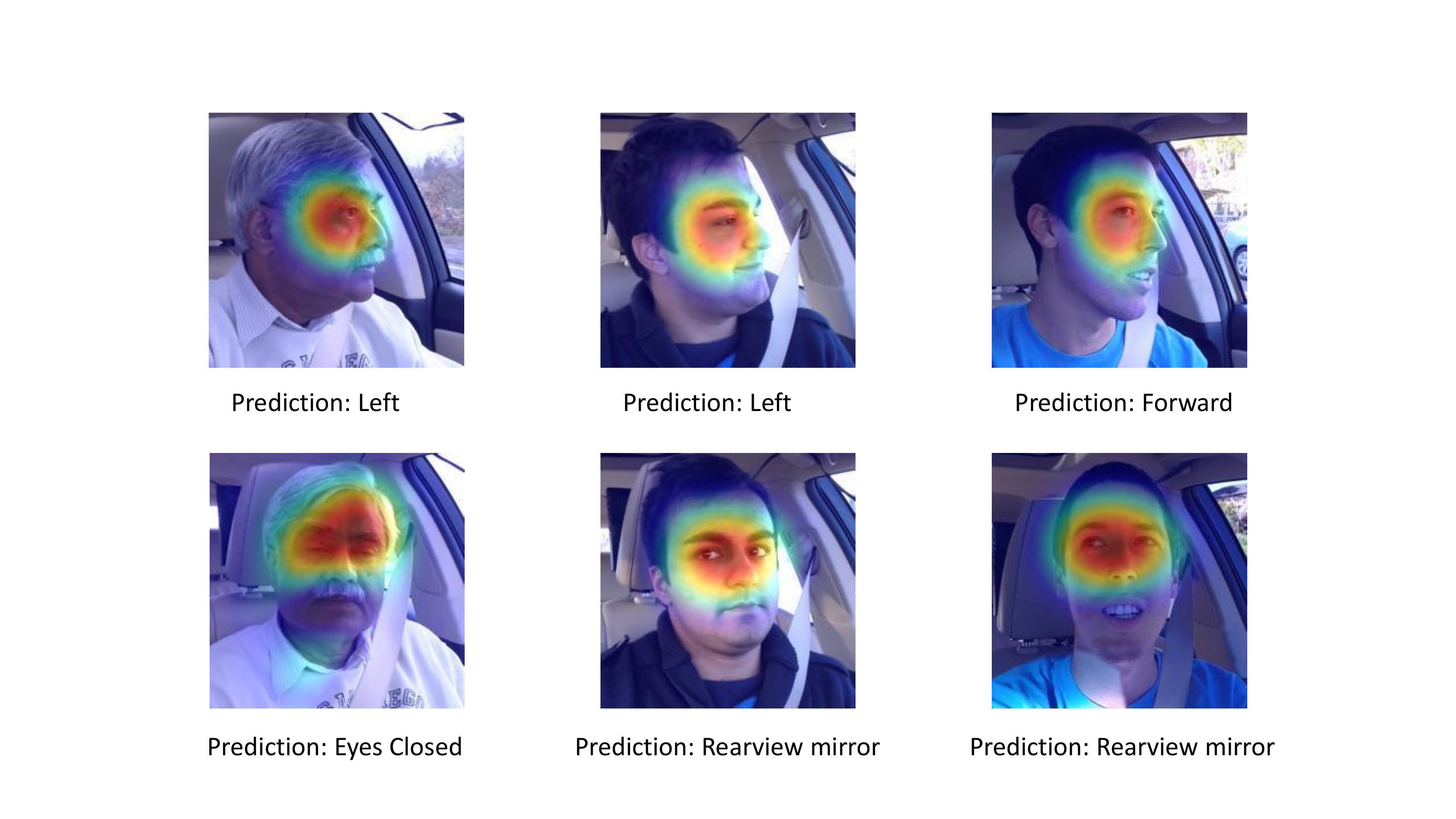}
      \caption{Class activation maps (CAMs) of the predicted class for different sample images. In the top 3 images, since the left eye of the driver is occluded by the face, our model learns to make a decision by looking at only one eye of the driver. In the bottom three images since both eyes are completely visible, our model makes a decision by looking at both eyes.}
      \label{focus}
\end{figure}

Second, the network also learns to intelligently focus on either one or both eyes of the driver. This can be observed in the activations of Image C vs the activations of Images A and B. In images A and B, the driver is looking at the radio and the rearview mirror and network uses both the eyes of the driver to make the decision. In Image C, the driver is looking at the speedometer and the network only uses the right eye of the driver to make the decision. The left eye is farther away from the camera and whenever the driver is looking to his left or his face is tilted, the left eye is self occluded by the face of the driver. This is further observed when we look at CAM of the predicted class for several different images in Fig. \ref{focus}. Thus, the network learns to deal with occlusion by intelligently focusing on either one eye or both eyes of the driver. 

Buoyed by the fact that the network learns to localize the eyes and observing much higher accuracies of the models trained on upper half of driver's face, we attempt to train our models on higher resolution Face Embedded FoV images. Since the SqueezeNet architecture does not contain any fully connected layers and only convolution layers, it can be finetuned on larger sized images. We believe that the model trained on upper half face images is able to extract finer features of the eye like the position and shape of iris and eyelid much better which explains it's better performance. Thus, increasing the resolution of Face Embedded FoV images should also help the model perform better. 

Table \ref{sq_res} shows the macro-average accuracies obtained by the network on training with higher resolution Face-embedded FoV images. The training settings were similar to what was described in \ref{training} and only the batch size was changed based on GPU memory constraints. It can be clearly observed that on increasing the resolution, the model starts performing much better. When the network was finetuned on $625\times625$ images, we achieve an accuracy of 92.13\%. Even though the performance is still lower than when the network is trained on upper half of face images, there is a huge advantage that no separate face detection step is required. Most modern state of the art object detectors consist of a region proposal network (RPN) and a detection network which further refines these proposals. These detectors are limited to perform real time at 30 fps. If we directly predict the gaze labels by skipping the face detection step, we only have to perform one forward pass through the network. This enables our system to perform real time. Further, the predictions won't be affected by inaccurate face detections.

\begin{table}[t!]
\centering
\caption{Performance of SqueezeNet architecture trained on Face embedded FoV images of varying resolutions}
\label{sq_res}
\renewcommand{\arraystretch}{2.4}
\begin{tabular}{|c|c|}
\hline
\textbf{Resolution} & \textbf{Macro-average accuracy} \\ \hline\hline
$224\times224$             & 89.37 \%                              \\ \hline
$448\times448$               & 90.78 \%                               \\ \hline
$625\times625$                & \textbf{92.13} \%                               \\ \hline
\end{tabular}
\end{table}

\subsection{Inference time for gaze estimation using different architectures}

We analyze the inference time of different CNN architectures used in this research study. The analysis was performed using Caffe's Matlab interface on a system with a Titan X GPU. Table \ref{runtime} lists the run time for a single forward pass of an image through various networks. As expected, the run time for AlexNet and SqueezeNet is much faster than VGG16 and Resnet50. Thus, finetuned SqueezeNet becomes the straightforward choice for gaze zone estimation because of its high performance (both in terms of speed and accuracy).

We see that our standalone system in Section \ref{noface}, finetuned SqueezeNet trained on $625\times625$ Face Embedded FoV images which achieves an accuracy of 92.13\%, comfortably runs in real time at 166.7 Hz. Our best performing model, finetuned SqueezeNet trained on upper half of the face, requires additional time for face detection. When using the face detector presented in \cite{yuen2016looking}, our system runs at 16 Hz. However, face detection is not the objective of this research study and the face detector used by us can be easily replaced by any other real time face detector or using a combination of detector and tracker.

\begin{table}[t!]
\centering
\caption{Inference times of the various CNNs used in this research study}
\label{runtime}
\renewcommand{\arraystretch}{1.4}
\begin{tabular}{|c|c|c|}
\hline
\textbf{CNN} & \textbf{Image resolution}  & \textbf{Run Time (ms)} \\ \hline\hline
AlexNet   & $227\times227$                   & 2.3                               \\ \hline
VGG16     & $224\times224$                   & 10                               \\ \hline
Resnet50  & $224\times224$                  & 17                               \\ \hline
SqueezeNet & $224\times224$                 & 2.5                           \\ \hline  
SqueezeNet & $448\times448$                 & 4                           \\ \hline  
SqueezeNet & $625\times625$                 & 6                           \\ \hline  

\end{tabular}
\end{table}

\section{Generalization on the Columbia Gaze Dataset}

In this section we test the generalization ability of our model on the Columbia Gaze Dataset\cite{smith2013gaze}. This dataset was created for sensing eye contact in an image. It has a total of 5,880 high resolution images of 56 subjects (32 males and 24 females) with extensive variability in the ethnicity of the subjects (21 Asians, 19 Whites, 8 South Asians, 7 Blacks and 4 Hispanics or Latinos). Further, 37 of the 56 subjects wore prescription glasses. \\

Subjects were seated at a distance of 2m from the camera and were asked to look at a grid of dots attached to a wall in front of them. For each subject, images were acquired for each combination of five horizontal head poses ($0^\circ$, $\pm5^\circ$, $\pm30^\circ$), seven horizontal gaze directions ($0^\circ$, $\pm5^\circ$, $\pm10^\circ$, $\pm±15^\circ$), and three vertical gaze directions ($0^\circ$, $\pm10^\circ$). Thus, there is a single image corresponding to a total of 105 pose-gaze configurations for each of the 56 subjects. \\

 As the problem (multiclass vs binary classification) and the dataset (Naturalistic driving data vs carefully collected data in a lab with a DSLR camera in perfect illumination conditions) are very different to what we have, we won't be comparing our method against theirs. Thus, instead of training a new network for this task, we run our best performing network on this dataset and attempt to analyze if our network can encode the different gaze directions on it. This should be possible as, on looking closely at the images of this dataset, we found that a few of the 105 pose-gaze configurations resemble the way we look forward (or towards other gaze zones) in the car. For each configuration, we check whether our network outputs a single gaze zone for majority of the subjects. We do so by plotting histograms as a bar graph where the y-axis represents the percentage of 56 subjects that output a particular gaze zone while the x-axis represents the gaze zones. We also calculate the normalized entropy for each configuration. Normalized entropy is defined as

\begin{equation}
  H_n(p) = -\sum_i \frac{p_i \log_b p_i}{\log_b n}
\end{equation}

where, $p_i$ is the fraction of subjects which output a particular gaze zone, $\textit{n } \text{is the number of classes and } H_n(p) \in [0, 1]. $ A low entropy indicates that the network successfully encodes the gaze direction.  

\begin{figure*}[t!]
      \centering
      \includegraphics[width=0.9\linewidth, height=15cm]{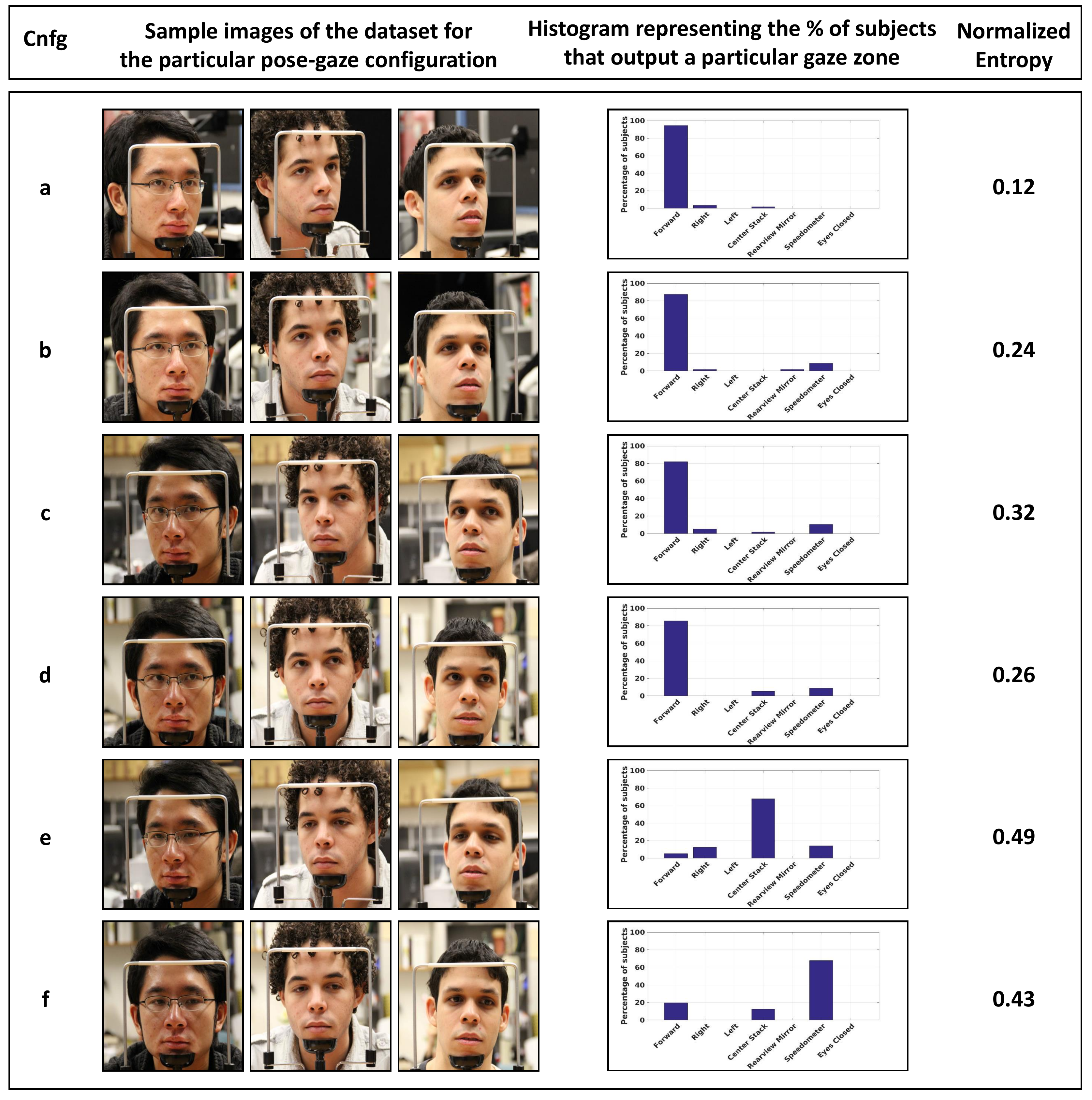}
      \caption{Sample images from 6 pose-gaze configurations of the Columbia Gaze dataset\cite{smith2013gaze}, the histograms of the predicted gaze zones by our best performing model on those configurations, and the normalized entropy. Our model successfully encodes the gaze direction on a completely different dataset with different camera pose, 56 new subjects of varying ethnicity with a large fraction of them wearing glasses. This shows the generalization ability of our model.}
      \label{columbia}
\end{figure*}

Fig \ref{columbia} contains sample images of the dataset for six carefully chosen configurations with varying head poses and gaze directions. These configurations resemble the way drivers look at different gaze zones in a car. Fig \ref{columbia} also contains the histogram and the normalized entropy values for each configurations. The first 4 rows of the figure contains the pose-gaze configurations in which 'Forward' was predicted as the gaze zone for majority of the subjects. This result makes intuitive sense when we have a closer look at the sample images of these configurations. In these images, the subjects are looking to the right of the camera, which is similar to the case of our naturalistic driving dataset. A closer look at configurations (a-d) also suggests that the network is not just encoding the head pose but also the gaze direction of the subjects. The head pose varies significantly in them but the subjects are still looking to the right of the camera and our network intuitively predicts 'forward'. 

Further, there were a total of 19 different configurations in which the subjects were looking to the right of the camera and the vertical gaze was $0^\circ$ or $10^\circ$, where our network predicts 'forward' as the gaze zone for more than 70\% of the subjects. When the subjects were looking to the right of the camera and the vertical gaze was $-10^\circ$, the network predicts 'Speedometer' as seen in configuration f of Fig \ref{columbia}. Similarly, when the subjects were looking to the left of the camera and the vertical gaze was $-10^\circ$, the network predicts 'Radio' as the gaze zone for majority of the subjects as seen in Fig \ref{columbia} configuration e. Again, looking closely at the sample images of the subjects in configurations e and f, these resemble very much the way drivers look at Radio and Speedometer with half open eyes. Finally, none of the configurations predicted 'Right', 'Left' as the majority gaze zone because the grid of the dots on which the subjects looked at in the Columbia Gaze Dataset only spanned $\pm±15^\circ$ in the horizontal direction. 'Eyes Closed' also wasn't predicted as the majority gaze zone as the dataset contains no images in which the eyes of the subjects are closed. 

These results suggest that our best performing model successfully encodes the gaze directions even on a completely new dataset without requiring any sort of training. This isn't straightforward because the camera pose in both the datasets are very different. In the Columbia gaze dataset, the camera was placed at eye level of the subject whereas in our naturalistic driving dataset, it is placed much above the eye level (just below the rearview mirror). The orientation of the camera with respect to the subject was also very different in both datasets. Further, the dataset contains 56 new subjects of various ethnicity with a large fraction of them also wearing prescription glasses. This shows the generalization ability of our model.

\section{Concluding Remarks}

Correct classification of driver's gaze is important as alerting the driver at the 'correct time' can prevent several road accidents. It will also help autonomous vehicles to determine driver distraction so as to calculate the appropriate \textit{handoff} time to the human driver. In literature, a large progress has been made towards personalized gaze zone estimation systems but not towards systems which can generalize to different drivers, cars, perspective and scale. Towards this end, we propose to use CNNs to classify driver's gaze into seven zones. The evaluations were made on a large naturalistic driving dataset (NDS) of 11 drives, driven by 10 subjects in 2 separate cars. Extensive ablation experiments were performed by evaluating the suitability of different CNN architectures and different input pre processing strategies for the gaze zone classification task. Four separate CNNs (AlexNet, VGG16, ResNet50 and SqueezeNet) were fine tuned on the collected NDS by training them on different image crop regions. It was found that a fine tuned SqueezeNet when trained on images of upper half of the face performs the best with an accuracy of 95.18\%. This is a large improvement over existing state of the art techniques for driver gaze zone classification. It was also shown that our network learns to localize the eyes of the driver without requiring any ground truth annotations of the eye or the face, thereby completely removing the need for face detection. Our standalone system which does not require any face detection, performs at an accuracy of 92.13\% while performing real time at 166.7 Hz on a GPU. Finally, we also showed that our best performing model successfully encodes the gaze directions on the diverse Columbia Gaze Dataset without requiring any training on it, thereby confirming its generalization capabilities. 

Future work in this direction will focus on adding more zones so as to obtain a finer estimate of driver's gaze. In the current implementation, the gaze zone predictions are made for each frame independently. In the future, we will also utilize temporal context using Long Short Term Memory (LSTM)\cite{hochreiter1997long}, which will help us capture the transitions from one gaze zone to another. The challenge with implementing an LSTM will however be to obtain continuous gaze zone image labels as opposed to labeled frames for discrete events separated across time.




\section*{Acknowledgments}

The authors would like to specially thank Sujitha Martin, Kevan Yuen and Nachiket Deo for their suggestions to improve this work. The authors also express our gratitude for all the valuable and constructive comments from the reviewers. The authors would also like to thank our sponsors and our colleagues at Laboratory for Intelligent and Safe Automobiles (LISA) for their massive help in data collection.

\bibliographystyle{IEEEtran}
\bibliography{root6}

\begin{thebibliography}{10}
\providecommand{\url}[1]{#1}
\csname url@rmstyle\endcsname
\providecommand{\newblock}{\relax}
\providecommand{\bibinfo}[2]{#2}
\providecommand\BIBentrySTDinterwordspacing{\spaceskip=0pt\relax}
\providecommand\BIBentryALTinterwordstretchfactor{4}
\providecommand\BIBentryALTinterwordspacing{\spaceskip=\fontdimen2\font plus
\BIBentryALTinterwordstretchfactor\fontdimen3\font minus
  \fontdimen4\font\relax}
\providecommand\BIBforeignlanguage[2]{{%
\expandafter\ifx\csname l@#1\endcsname\relax
\typeout{** WARNING: IEEEtran.bst: No hyphenation pattern has been}%
\typeout{** loaded for the language `#1'. Using the pattern for}%
\typeout{** the default language instead.}%
\else
\language=\csname l@#1\endcsname
\fi
#2}}

\bibitem{eriksson2016take}
A.~Eriksson and N.~Stanton, ``Take-over time in highly automated vehicles:
  non-critical transitions to and from manual control,'' \emph{Human Factors},
  2016.

\bibitem{fitch2013impact}
G.~M. Fitch, S.~A. Soccolich, F.~Guo, J.~McClafferty, Y.~Fang, R.~L. Olson,
  M.~A. Perez, R.~J. Hanowski, J.~M. Hankey, and T.~A. Dingus, ``The impact of
  hand-held and hands-free cell phone use on driving performance and
  safety-critical event risk,'' Tech. Rep., 2013.

\bibitem{rueda2004influence}
T.~Rueda-Domingo, P.~Lardelli-Claret, J.~de~Dios Luna-del Castillo, J.~J.
  Jim{\'e}nez-Mole{\'o}n, M.~Garc{\i}́a-Mart{\i}́n, and A.~Bueno-Cavanillas,
  ``The influence of passengers on the risk of the driver causing a car
  collision in spain: Analysis of collisions from 1990 to 1999,''
  \emph{Accident Analysis \& Prevention}, vol.~36, no.~3, pp. 481--489, 2004.

\bibitem{braitman2014effect}
K.~A. Braitman, N.~K. Chaudhary, and A.~T. McCartt, ``Effect of passenger
  presence on older drivers’ risk of fatal crash involvement,'' \emph{Traffic
  injury prevention}, vol.~15, no.~5, pp. 451--456, 2014.

\bibitem{li2016detecting}
N.~Li and C.~Busso, ``Detecting drivers' mirror-checking actions and its
  application to maneuver and secondary task recognition,'' \emph{IEEE
  Transactions on Intelligent Transportation Systems}, vol.~17, no.~4, pp.
  980--992, 2016.

\bibitem{ahlstrom2013gaze}
C.~Ahlstrom, K.~Kircher, and A.~Kircher, ``A gaze-based driver distraction
  warning system and its effect on visual behavior,'' \emph{IEEE Transactions
  on Intelligent Transportation Systems}, vol.~14, no.~2, pp. 965--973, 2013.

\bibitem{doshi2011tactical}
A.~Doshi and M.~M. Trivedi, ``Tactical driver behavior prediction and intent
  inference: A review,'' in \emph{Intelligent Transportation Systems (ITSC),
  2011 14th International IEEE Conference on}.\hskip 1em plus 0.5em minus
  0.4em\relax IEEE, 2011, pp. 1892--1897.

\bibitem{martin2017behavior}
S.~Martin and M.~M. Trivedi, ``Gaze fixations and dynamics for behavior
  modeling and prediction of on-road driving maneuvers,'' in \emph{Intelligent
  Vehicles Symposium Proceedings, 2017 IEEE}.\hskip 1em plus 0.5em minus
  0.4em\relax IEEE, 2017.

\bibitem{tawari2014robust}
A.~Tawari and M.~M. Trivedi, ``Robust and continuous estimation of driver gaze
  zone by dynamic analysis of multiple face videos,'' in \emph{Intelligent
  Vehicles Symposium Proceedings, 2014 IEEE}.\hskip 1em plus 0.5em minus
  0.4em\relax IEEE, 2014, pp. 344--349.

\bibitem{tawari2014driver}
A.~Tawari, K.~H. Chen, and M.~M. Trivedi, ``Where is the driver looking:
  Analysis of head, eye and iris for robust gaze zone estimation,'' in
  \emph{Intelligent Transportation Systems (ITSC), 2014 IEEE 17th International
  Conference on}.\hskip 1em plus 0.5em minus 0.4em\relax IEEE, 2014, pp.
  988--994.

\bibitem{vasli2016driver}
B.~Vasli, S.~Martin, and M.~M. Trivedi, ``On driver gaze estimation:
  Explorations and fusion of geometric and data driven approaches,'' in
  \emph{Intelligent Transportation Systems (ITSC), 2016 IEEE 19th International
  Conference on}.\hskip 1em plus 0.5em minus 0.4em\relax IEEE, 2016, pp.
  655--660.

\bibitem{fridman2015driver}
L.~Fridman, P.~Langhans, J.~Lee, and B.~Reimer, ``Driver gaze region estimation
  without using eye movement,'' \emph{arXiv preprint arXiv:1507.04760}, 2015.

\bibitem{fridman2016owl}
L.~Fridman, J.~Lee, B.~Reimer, and T.~Victor, ``Owl and lizard: patterns of
  head pose and eye pose in driver gaze classification,'' \emph{IET Computer
  Vision}, vol.~10, no.~4, pp. 308--313, 2016.

\bibitem{choi2016real}
I.-H. Choi, S.~K. Hong, and Y.-G. Kim, ``Real-time categorization of driver's
  gaze zone using the deep learning techniques,'' in \emph{Big Data and Smart
  Computing (BigComp), 2016 International Conference on}.\hskip 1em plus 0.5em
  minus 0.4em\relax IEEE, 2016, pp. 143--148.

\bibitem{oquab2014learning}
M.~Oquab, L.~Bottou, I.~Laptev, and J.~Sivic, ``Learning and transferring
  mid-level image representations using convolutional neural networks,'' in
  \emph{Proceedings of the IEEE conference on computer vision and pattern
  recognition}, 2014, pp. 1717--1724.

\bibitem{deng2009imagenet}
J.~Deng, W.~Dong, R.~Socher, L.-J. Li, K.~Li, and L.~Fei-Fei, ``Imagenet: A
  large-scale hierarchical image database,'' in \emph{Computer Vision and
  Pattern Recognition, 2009. CVPR 2009. IEEE Conference on}.\hskip 1em plus
  0.5em minus 0.4em\relax IEEE, 2009, pp. 248--255.

\bibitem{dong2011driver}
Y.~Dong, Z.~Hu, K.~Uchimura, and N.~Murayama, ``Driver inattention monitoring
  system for intelligent vehicles: A review,'' \emph{IEEE transactions on
  intelligent transportation systems}, vol.~12, no.~2, pp. 596--614, 2011.

\bibitem{bergasa2006real}
L.~M. Bergasa, J.~Nuevo, M.~A. Sotelo, R.~Barea, and M.~E. Lopez, ``Real-time
  system for monitoring driver vigilance,'' \emph{IEEE Transactions on
  Intelligent Transportation Systems}, vol.~7, no.~1, pp. 63--77, 2006.

\bibitem{ji2001real}
Q.~Ji and X.~Yang, ``Real time visual cues extraction for monitoring driver
  vigilance,'' in \emph{International Conference on Computer Vision
  Systems}.\hskip 1em plus 0.5em minus 0.4em\relax Springer, 2001, pp.
  107--124.

\bibitem{ji2002real}
Q.~Ji and X.~Yang, ``Real-time eye, gaze, and face pose tracking for monitoring
  driver vigilance,'' \emph{Real-Time Imaging}, vol.~8, no.~5, pp. 357--377,
  2002.

\bibitem{morimoto2000pupil}
C.~H. Morimoto, D.~Koons, A.~Amir, and M.~Flickner, ``Pupil detection and
  tracking using multiple light sources,'' \emph{Image and vision computing},
  vol.~18, no.~4, pp. 331--335, 2000.

\bibitem{lee2011real}
S.~J. Lee, J.~Jo, H.~G. Jung, K.~R. Park, and J.~Kim, ``Real-time gaze
  estimator based on driver's head orientation for forward collision warning
  system,'' \emph{IEEE Transactions on Intelligent Transportation Systems},
  vol.~12, no.~1, pp. 254--267, 2011.

\bibitem{ishikawa2004passive}
T.~Ishikawa, ``Passive driver gaze tracking with active appearance models,''
  2004.

\bibitem{smith2003determining}
P.~Smith, M.~Shah, and N.~da~Vitoria~Lobo, ``Determining driver visual
  attention with one camera,'' \emph{IEEE transactions on intelligent
  transportation systems}, vol.~4, no.~4, pp. 205--218, 2003.

\bibitem{murphy2009head}
E.~Murphy-Chutorian and M.~M. Trivedi, ``Head pose estimation in computer
  vision: A survey,'' \emph{IEEE transactions on pattern analysis and machine
  intelligence}, vol.~31, no.~4, pp. 607--626, 2009.

\bibitem{he2016deep}
K.~He, X.~Zhang, S.~Ren, and J.~Sun, ``Deep residual learning for image
  recognition,'' in \emph{Proceedings of the IEEE Conference on Computer Vision
  and Pattern Recognition}, 2016, pp. 770--778.

\bibitem{Simonyan14c}
K.~Simonyan and A.~Zisserman, ``Very deep convolutional networks for
  large-scale image recognition,'' \emph{CoRR}, vol. abs/1409.1556, 2014.

\bibitem{krizhevsky2012imagenet}
A.~Krizhevsky, I.~Sutskever, and G.~E. Hinton, ``Imagenet classification with
  deep convolutional neural networks,'' in \emph{Advances in neural information
  processing systems}, 2012, pp. 1097--1105.

\bibitem{iandola2016squeezenet}
F.~N. Iandola, S.~Han, M.~W. Moskewicz, K.~Ashraf, W.~J. Dally, and K.~Keutzer,
  ``Squeezenet: Alexnet-level accuracy with 50x fewer parameters and< 0.5 mb
  model size,'' \emph{arXiv preprint arXiv:1602.07360}, 2016.

\bibitem{yuen2016looking}
K.~Yuen, S.~Martin, and M.~M. Trivedi, ``Looking at faces in a vehicle: A deep
  cnn based approach and evaluation,'' in \emph{Intelligent Transportation
  Systems (ITSC), 2016 IEEE 19th International Conference on}.\hskip 1em plus
  0.5em minus 0.4em\relax IEEE, 2016, pp. 649--654.

\bibitem{he2015delving}
K.~He, X.~Zhang, S.~Ren, and J.~Sun, ``Delving deep into rectifiers: Surpassing
  human-level performance on imagenet classification,'' in \emph{Proceedings of
  the IEEE international conference on computer vision}, 2015, pp. 1026--1034.

\bibitem{kingma2014adam}
D.~Kingma and J.~Ba, ``Adam: A method for stochastic optimization,''
  \emph{arXiv preprint arXiv:1412.6980}, 2014.

\bibitem{jia2014caffe}
Y.~Jia, E.~Shelhamer, J.~Donahue, S.~Karayev, J.~Long, R.~Girshick,
  S.~Guadarrama, and T.~Darrell, ``Caffe: Convolutional architecture for fast
  feature embedding,'' \emph{arXiv preprint arXiv:1408.5093}, 2014.

\bibitem{martin2016vision}
S.~Martin, K.~Yuen, and M.~M. Trivedi, ``Vision for intelligent vehicles \&
  applications (viva): Face detection and head pose challenge,'' in
  \emph{Intelligent Vehicles Symposium (IV), 2016 IEEE}.\hskip 1em plus 0.5em
  minus 0.4em\relax IEEE, 2016, pp. 1010--1014.

\bibitem{zhou2016learning}
B.~Zhou, A.~Khosla, A.~Lapedriza, A.~Oliva, and A.~Torralba, ``Learning deep
  features for discriminative localization,'' in \emph{Proceedings of the IEEE
  Conference on Computer Vision and Pattern Recognition}, 2016, pp. 2921--2929.

\bibitem{smith2013gaze}
B.~A. Smith, Q.~Yin, S.~K. Feiner, and S.~K. Nayar, ``Gaze locking: passive eye
  contact detection for human-object interaction,'' in \emph{Proceedings of the
  26th annual ACM symposium on User interface software and technology}.\hskip
  1em plus 0.5em minus 0.4em\relax ACM, 2013, pp. 271--280.

\bibitem{hochreiter1997long}
S.~Hochreiter and J.~Schmidhuber, ``Long short-term memory,'' \emph{Neural
  computation}, vol.~9, no.~8, pp. 1735--1780, 1997.

\end{thebibliography}

\begin{IEEEbiography}[{\includegraphics[width=1in,height=1.25in,clip,keepaspectratio]{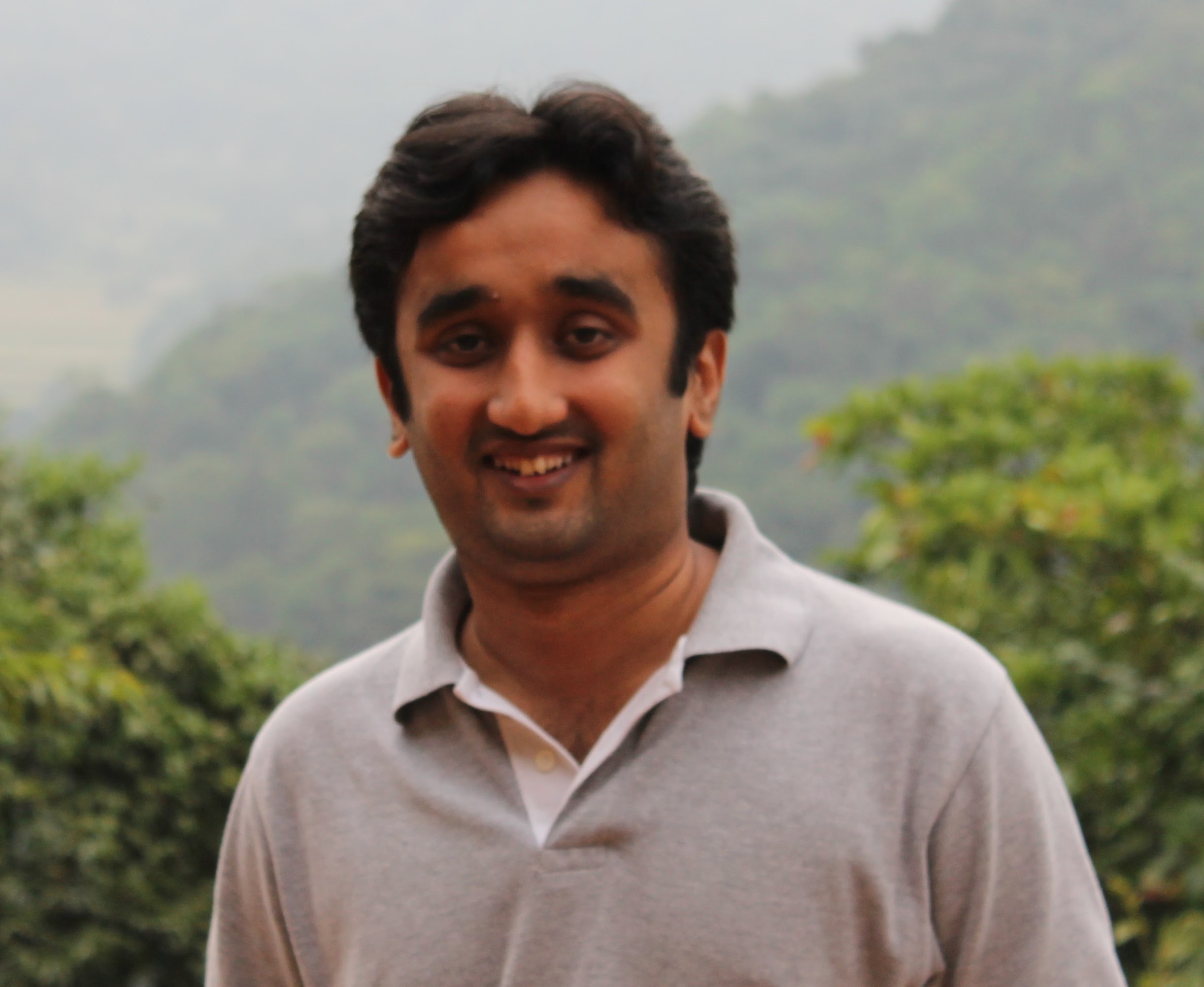}}]{Sourabh Vora}
received his BS degree in Electronics and Communications Engineering (ECE) from Birla Institute of Technology and Science (BITS) Pilani - Hyderabad Campus. He received his MS degree in Electrical and Computer Engineering (ECE) from University of California, San Diego (UCSD) where he was associated with the Computer Vision and Robotics Research (CVRR) Lab. His research interests lie in the field of Computer Vision and Machine Learning. He is currently working as a Computer Vision Engineer at nuTonomy, Santa Monica.
\end{IEEEbiography}

\begin{IEEEbiography}[{\includegraphics[width=1in,height=1.25in,clip,keepaspectratio]{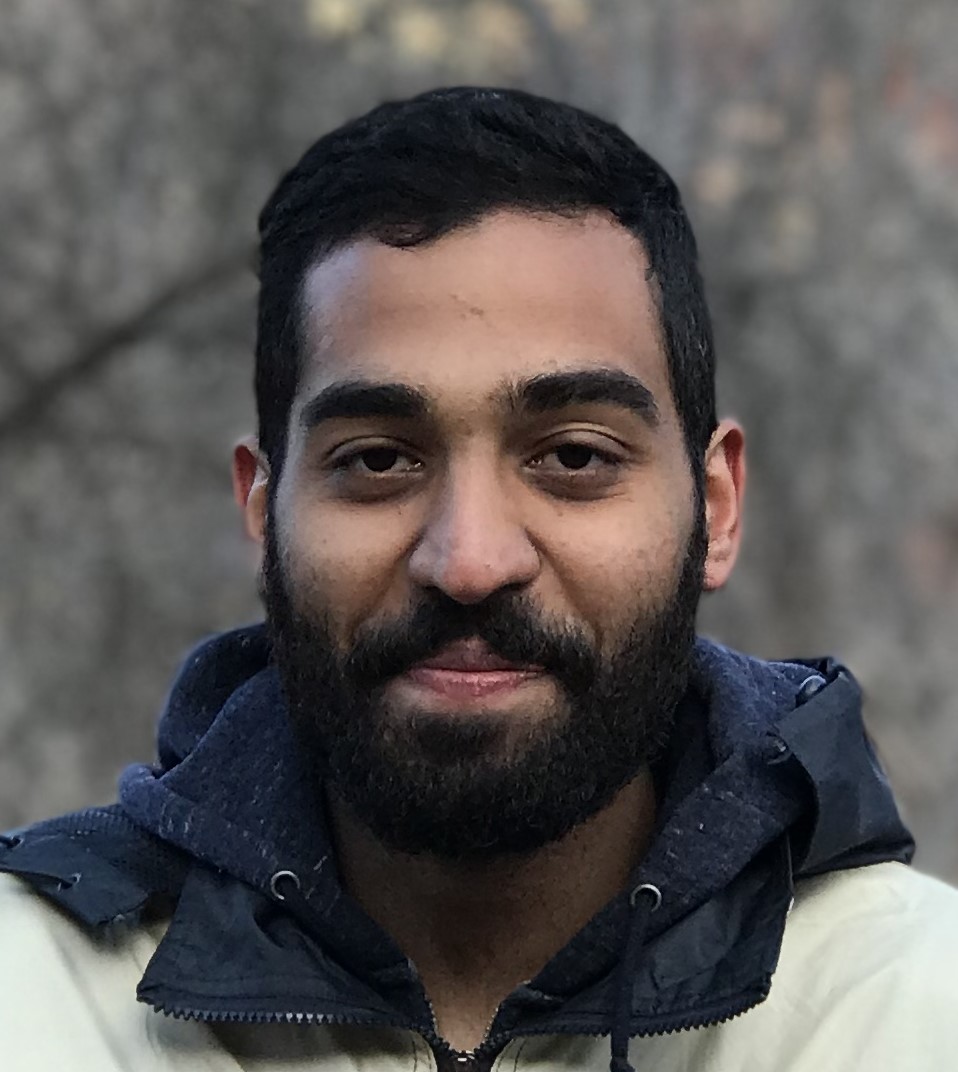}}]{Akshay Rangesh}
is currently working towards his PhD in electrical engineering from the University of California at San Diego (UCSD), with a focus on intelligent systems, robotics, and control. His research interests span computer vision and machine learning, with a focus on object detection and tracking, human activity recognition, and driver safety systems in general. He is also particularly interested in sensor fusion and multi-modal approaches for real time algorithms.
\end{IEEEbiography}

\begin{IEEEbiography}[{\includegraphics[width=1in,height=1.25in,clip,keepaspectratio]{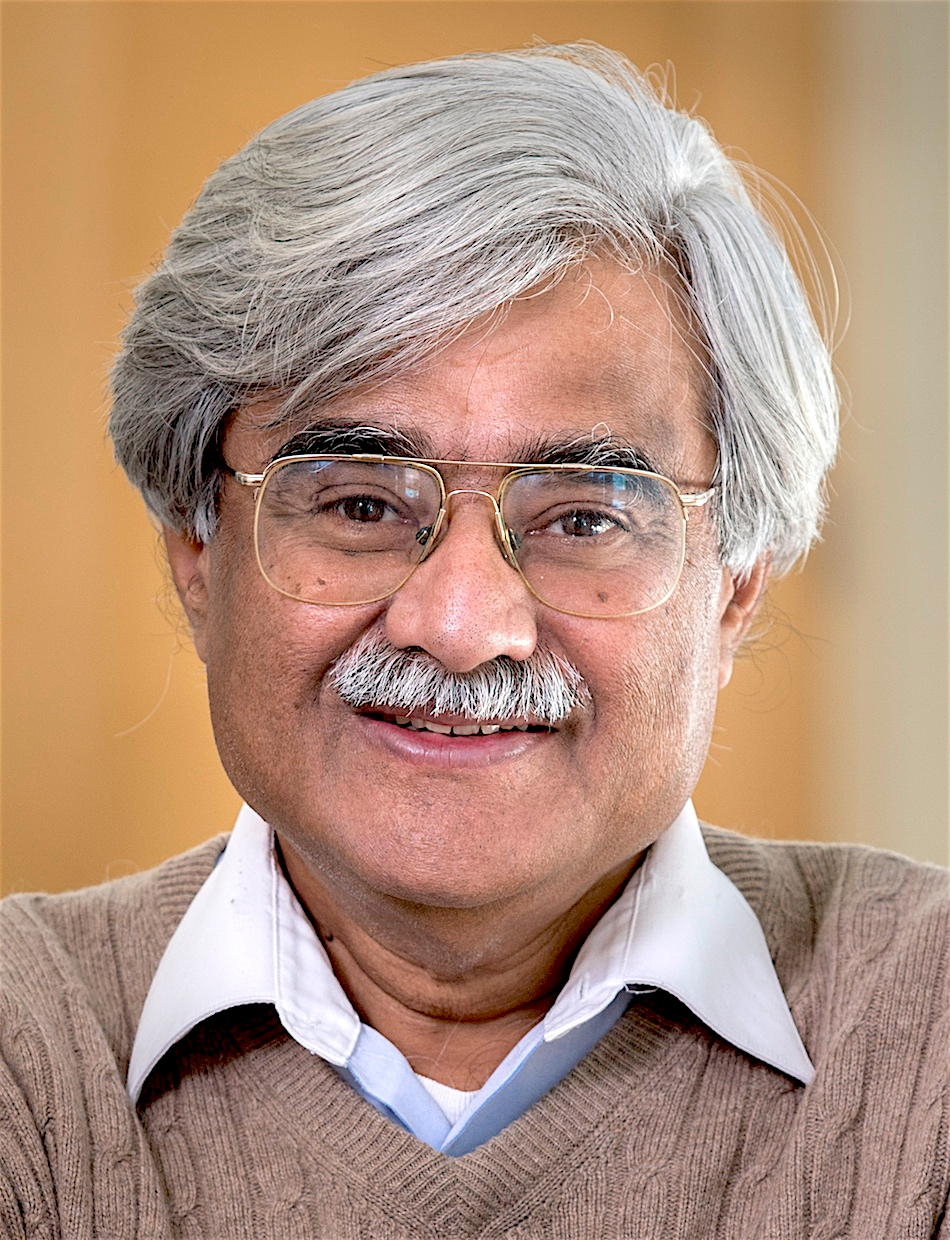}}]{Mohan Manubhai Trivedi}
is a Distinguished Professor at University of California, San Diego (UCSD) and the founding director of the UCSD LISA: Laboratory for Intelligent and Safe Automobiles,
winner of the IEEE ITSS Lead Institution Award (2015). Currently, Trivedi and his team
are pursuing research in intelligent vehicles, machine perception, machine learning, human-robot interactivity, driver assistance, active safety systems. Three of his students have received "best dissertation" recognitions. Trivedi is a Fellow of IEEE, ICPR and SPIE. He received the IEEE ITS Society's highest accolade "Outstanding Research Award" in 2013. Trivedi serves frequently as a consultant to industry and government agencies in the USA and abroad. 
\end{IEEEbiography}

\end{document}